%% file: KDD/sample-sigconf.tex
\documentclass[sigconf]{acmart}
\AtBeginDocument{%
  }

\copyrightyear{2026}
\acmYear{2026}
\setcopyright{cc}
\setcctype{by}
\acmConference[KDD 2026] {Proceedings of the 32nd ACM SIGKDD Conference on Knowledge Discovery and Data Mining V.2}{August 9--13, 2026}{Jeju Island, Republic of Korea.}
\acmBooktitle{Proceedings of the 32nd ACM SIGKDD Conference on Knowledge Discovery and Data Mining V.2 (KDD 2026), August 9--13, 2026, Jeju Island, Republic of Korea}
\acmISBN{979-8-4007-2259-2/2026/08} 
\acmDOI{https://doi.org/10.1145/3770855.3817754}

\usepackage{graphicx}

\newtheorem{theorem}{Theorem}[section]
\newtheorem{proposition}[theorem]{Proposition}

\newtheorem{definition}[theorem]{Definition}

\usepackage{multirow}
\usepackage{enumitem}

\usepackage{graphicx}
\usepackage{makecell}
\usepackage{algorithm}
\usepackage{algorithmic}

\usepackage{amsfonts}       
\usepackage{nicefrac}       
\usepackage{microtype}      
\usepackage{xcolor}         
\newcommand{\R}{\mathcal{R}}%
\newcommand{\vs}{\mathbf{s}}%
\newcommand{\mS}{\mathcal{S}}%
\DeclareMathOperator{\diag}{{\rm diag}}
\begin{document}

\title{Efficient and Scalable Neural-Symbolic Search for\\
 Complex Query Answering over Incomplete Knowledge Graphs }

\author{Weizhi Fei}
\orcid{0009-0000-1709-1605}
\affiliation{%
  \institution{Department of Mathematical Sciences, Tsinghua University}
  \city{Beijing}
  \state{}
  \country{China}
}
\email{fwz22@mails.tsinghua.edu.cn}

\author{Zihao Wang}
\orcid{0000-0002-3919-0396}
\authornote{Corresponding author.}
\affiliation{
  \institution{Department of Computer Science and Engineering, Hong Kong University of Science and Technology}
  \city{Hong Kong}
  \state{}
  \country{Hong Kong}
}
\email{zwanggc@connect.ust.hk}

\author{Hang Yin}
\orcid{0009-0001-5264-1630}
\affiliation{
  \institution{Squarepoint Capital}
  \city{Singapore}
  \state{}
  \country{Singapore}
}
\email{yinhwx@gmail.com}

\author{Shukai Zhao}
\orcid{0009-0001-4689-4826}
\affiliation{
  \institution{Department of Computer Sciences, University of Rochester}
  \city{New York}
  \state{}
  \country{USA}
}
\email{szhao27@ur.rochester.edu}

\author{Wei Zhang}
\orcid{0009-0004-6255-8373}
\affiliation{
  \institution{Department of Mathematical Sciences, Tsinghua University}
  \city{Beijing}
  \state{}
  \country{China}
}
\email{zhangwei.2020@tsinghua.org.cn}

\author{Yangqiu Song}
\orcid{0000-0002-7818-6090}
\affiliation{%
\institution{Department of Computer Science and Engineering, Hong Kong University of Science and Technology}
  \city{Hong Kong}
  \state{}
  \country{Hong Kong}
}
\email{yqsong@cse.ust.hk}


\begin{abstract}
Complex Query Answering (CQA) is a crucial reasoning task over Knowledge Graphs (KGs), which aims to answer first-order logical queries from incomplete KGs. While existing neural-symbolic methods achieve strong performance, they face significant complexity bottlenecks: quadratic data complexity scaling with the number of entities, and NP-hard query complexity for cyclic queries. Consequently, these approaches struggle to scale effectively to large knowledge graphs and complex queries. To address these limitations, we propose an efficient and scalable symbolic search method comprising two key components: (1) constraint strategies that drastically reduce the variable search domain, lowering data complexity; and (2) a local search algorithm that approximately solves NP-hard cyclic queries. Experiments on various CQA benchmarks demonstrate that, for tree-form queries, our method achieves 97\% relative MRR with a 10× speedup using only 10\% of the search space. Furthermore, it demonstrates robust performance on complex cyclic queries and large-scale KGs, effectively alleviating efficiency and scalability challenges. Our code is provided in \url{https://github.com/HKUST-KnowComp/NLISA_KDD2026}.
\end{abstract}




\maketitle
\newcommand\kddavailabilityurl{https://zenodo.org/records/20355431}
\ifdefempty{\kddavailabilityurl}{}{
\begingroup\small\noindent\raggedright\textbf{Resource Availability:}\\
The source code of this paper has been made publicly available at \url{\kddavailabilityurl}.
\endgroup
}

\input{cc_codex/NLISA_arxiv/NLISA_main}

\section{Acknowledgments}
The authors of this paper were supported by the National Key Research and Development Program of China (2025YFE0200500), the ITSP Platform Research Project (ITS/189/23FP) from ITC of Hong Kong, SAR, China, and the AoE (AoE/E-601/24-N), the CRF (No. C6004-25G), the RIF (R6021-20), and the GRF (16205322) from RGC of Hong Kong, SAR, China.



\bibliographystyle{ACM-Reference-Format}
\balance
\bibliography{my_refs}

\appendix

\input{cc_codex/NLISA_arxiv/NLISA_appendix}
\end{document}

%% file: cc_codex/NLISA_arxiv/NLISA_main.tex
\section{Introduction}
\begin{figure*}[t]
    \centering
    \includegraphics[width=\linewidth]{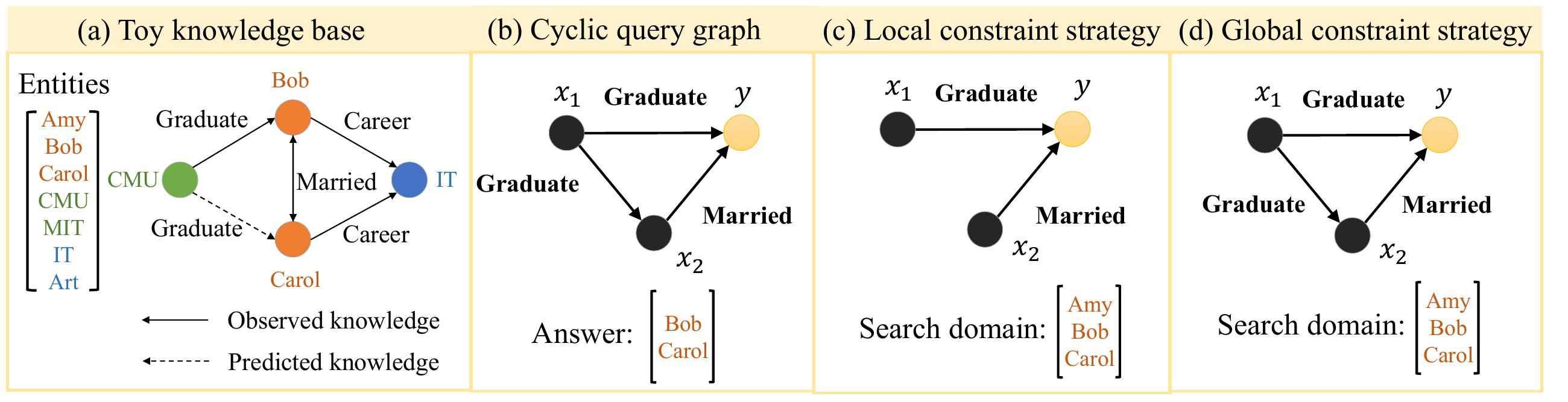}
    \caption{ \textbf{(a)} One toy knowledge graph recording  a simple social network.  \textbf{(b)} A simple cyclic query, with formula $\exists x_1,x_2, \text{Graduate}(x_1, x_2) \land  \text{Graduate}(x_1, y) \land  \text{Married}(x_2, y)$. This can be interpreted as ``Find a person who graduated from the same institution (s) as his/her spouse''. Since the answers $\{\text{Bob}, \text{Carol}\}$ must be inferred with predicted edges. To predict the potential answers, CQA models must consider all entities within the given KG.  \textbf{(c)} and \textbf{(d)} present the used subgraph for free variable $y$ under the local constraints strategy and the global constraints strategy, respectively. The toy KG illustrates only the constraints used by the example; the pruned domains are produced by neural scores over candidate entities. We then utilize the pruned domain from constraints strategies to accelerate the search process.}\label{fig:query graph and constraints}
\end{figure*}

Complex Query Answering (CQA)~\citep{wang_logical_2022,ren_neural_2023} aims to answer complex logical queries over incomplete knowledge graphs (KGs). Although existing KGs are powerful knowledge bases for encoding real-world facts, they inevitably suffer from incompleteness~\citep{yang_rethinking_2022,peng2023knowledge}. Given an observed incomplete KG, CQA aims to infer missing answers to complex logical queries~\citep{ren_beta_2020} by exploiting the generalization capability of machine learning (ML) models. The queries considered in CQA are typically formulated as first-order logic (FOL) queries with existential quantification, which capture complex relational structures among entities. We provide an example as illustrated in Figure~\ref{fig:query graph and constraints}.

Traditional graph traversal methods lack generalization ability and therefore often fail to recover implicit knowledge that can be inferred from observed facts~\citep{hamilton_embedding_2018}. Meanwhile, large language models face challenges related to hallucination~\citep{rawte2023survey} and struggle to precisely memorize and utilize factual knowledge, particularly domain-specific knowledge~\citep{zheng2024clr}. To address these challenges, CQA provides a principled framework that effectively combines logical faithfulness with neural generalization. As a result, it enables a wide range of downstream applications, including recommendation and query-answering systems~\citep{bai2024understanding,ren_neural_2023}.

With recent advances, numerous ML models have been proposed for CQA~\citep{ren_neural_2023}. Among them, neural-symbolic search methods~\citep{amayuelas_neural_2021,yin_rethinking_2023} have emerged as a promising paradigm, offering faithful, interpretable reasoning while achieving state-of-the-art (SoTA) performance. However, their computational complexity still poses significant scalability challenges.

\begin{table}[t]
    \centering
    \caption{The complexity and performance (MRR) of different neural-symbolic search methods. The symbol * indicates tree decomposition (approximation) for cycles, and $\kappa$ denotes the search-domain reduction factor introduced by our method. For tree-form queries, the worst-case asymptotic order remains quadratic if $\kappa$ is treated as a constant; the gain comes from the reduced effective domain size. }\label{tab: complexity and perform}
    \resizebox{0.48\textwidth}{!}{
    \begin{tabular}{ccccc}
    \toprule
         & \multicolumn{2}{c}{Complexity} &   \multicolumn{2}{c}{Performance}  \\ \midrule
        Query category  & Tree & Cyclic & Tree & Cyclic \\ \midrule
        QTO~\citep{bai_answering_2023} & $\mathcal{O}(n|\mathcal{E}|^2)$ & $\mathcal{O}(n|\mathcal{E}|^2)$* & 37.1 & 36.8 \\ 
        FIT~\citep{yin_rethinking_2023} & $\mathcal{O}(n|\mathcal{E}|^2)$ &$\mathcal{O}(|\mathcal{E}|^n)$ & 37.1 & 42.0 \\ 
        NLISA (our) & $\mathcal{O}(n(\frac{|\mathcal{E}|}{\kappa})^2)$ & $\mathcal{O}(n(\frac{|\mathcal{E}|}{\kappa})^2)$ & 37.3 & 40.5 \\ \bottomrule
    \end{tabular}}
    \vspace{-1.5em}
\end{table}

We provide a detailed complexity analysis in Table~\ref{tab: complexity and perform} to show the existing complexity issues. Let $|\mathcal{E}|$ be the size of the entities within KGs and $n$ be the query size. For simple tree-form queries, the complexity of QTO and FIT is $\mathcal{O}(n|\mathcal{E}|^2)$. From the perspective of data complexity, this complexity grows quadratically with the size of the entities, making it challenging to scale to large-scale KGs~\citep{ren_smore_2022}. For cyclic queries, the complexity of FIT is even $\mathcal{O}(|\mathcal{E}|^n)$. From the perspective of query complexity, this complexity is NP-hard and increases exponentially with query size. This complexity bottleneck limits the efficiency in handling complex cyclic queries~\citep{yin_rethinking_2023}. To improve both scalability and efficiency, we propose NLISA, which maintains comparable performance while substantially reducing the effective search domain and avoiding exponential enumeration for cyclic queries, as shown in Table~\ref{tab: complexity and perform}.

Our first technical contribution is the proposed Neural Logical Indices (NLI) to narrow the search domain for involved variables, thereby reducing the practical cost of symbolic search. This technique is motivated by well-studied Constraint Satisfaction Problems (CSPs)~\citep{Chen2011ArcCA,gottlob_comparison_2000}, and the sparsity nature observed in KGs~\citep{pujara2017sparsity}. In Appendix~\ref{app:theory}, we provide an explanatory probabilistic analysis for why correct answers can be retained with high probability under mild modeling assumptions. Furthermore, we provide two strategies to flexibly compute NLI, including the \textbf{local constraints} involving neighborhood constraints  and \textbf{global constraints} strategy using global constraints. More details are presented in Section~\ref{Sec: neural logical index}. 


The second technical contribution is an approximate search algorithm to address \textbf{NP-hard} cyclic queries with quadratic complexity in the pruned domains. Our approximate solution autoregressively searches the most likely assignments for variables within cyclic structures and employs local search to avoid exponential growth of the search space. Notably, this algorithm supports parallel execution, significantly enhancing efficiency. Detailed introduction is shown in Section~\ref{sec: effcient search}. 


To validate the effectiveness of our method, we conduct comprehensive experiments on classic CQA benchmarks. Our experimental results reveal three key insights. (1) The search space for variables can be significantly reduced, achieving $97\%$ MRR with a $10 \times$ speedup. (2) For cyclic queries, our method achieves $50 \times$ speedup while maintaining $95\%$ MRR. (3) We demonstrate that our method can be extended to the KG with 400,000 entities, where prior neural-symbolic solvers run out of memory, highlighting the practical scalability of our method.


\section{Related work}
Complex Query Answering (CQA) over incomplete knowledge graphs (KGs) is a fundamental task in knowledge representation and reasoning, requiring both knowledge discovery and complex logical inference. The goal of CQA is to enhance the expressiveness of supported query languages while accurately inferring unseen answers from incomplete observations. Over time, CQA has progressively evolved by expanding the scope of supported logical queries, ranging from conjunctive queries with directed acyclic graph (DAG) structures~\citep{hamilton_embedding_2018} to general existential first-order queries that may contain cyclic structures~\citep{yin_rethinking_2023}.

Existing CQA approaches can be broadly categorized into two lines of research. The first line consists of \textbf{query embedding methods}, which embed logical queries and entities into a shared latent space and retrieve answers based on distance or similarity functions. In these methods, logical operations are mapped to differentiable set operations organized in an operator tree~\citep{wang_benchmarking_2021,ren_neural_2023} and are implemented via neural networks to approximate their semantics. To induce appropriate inductive biases, query embedding methods adopt expressive geometric or probabilistic representations, such as vectors, boxes, and beta distributions~\citep{hamilton_embedding_2018,ren_query2box_2020,ren_beta_2020}. Although their representational capacity has been extensively studied~\citep{zhang_cone_2021,choudhary_probabilistic_2021}, query embedding methods still suffer from limitations in both inference accuracy and logical expressiveness~\citep{yin_rethinking_2023}. Recent query embedding methods such as Query2Triple and context-aware query representation learning further improve tree-structured query answering, but they are not designed for cyclic EFO1 queries. In a related vein, purely neural approaches directly predict the probability of each candidate entity as an answer in an end-to-end manner, typically using transformers or graph neural networks to capture multi-hop dependencies and structural patterns. Compared with these purely neural models, neural-symbolic search keeps the reasoning process more traceable through domain construction, variable assignment, and constraint checking.

The second line comprises \textbf{neural-symbolic search} methods~\citep{arakelyan_complex_2020,bai_answering_2023,yin_rethinking_2023}, which employ knowledge graph embedding models~\citep{bordes_translating_2013} as neural backbones to predict missing facts and adopt fuzzy logic~\citep{klir_fuzzy_1995} to model logical operators. Representative methods such as QTO~\citep{bai_answering_2023} and FIT~\citep{yin_rethinking_2023} materialize completed score matrices of size $\mathcal{O}(|\mathcal{R}||\mathcal{E}|^2)$ and perform symbolic-style search under fuzzy semantics. QTO supports exact inference for tree-structured queries, while FIT extends this framework to general existential first-order queries by enumerating variable assignments to handle cyclic structures, resulting in a worst-case complexity of $\mathcal{O}(|\mathcal{E}|^n)$~\citep{yin_rethinking_2023}. In contrast, our proposed method, NLISA, reduces the effective domain searched by neural-symbolic methods and replaces exact cyclic enumeration with approximate local optimization, thereby improving their practical scalability.

\section{Background}
\subsection{Preliminary}
To simplify the discussion, we define the EFO1 query with the Disjunctive Normal Form (DNF).
\begin{definition}[Knowledge graph]
    Let $\mathcal{E}$ and $\mathcal{R}$ be the finite set of entities and relations, a knowledge graph is a collection of factual triples $\mathcal{G} = \{(s_i, r_i, o_i)\}$, where $s_i$ is the subject entity, $o_i$ is the object entity, and $r_i$ is a relation predicate.
\end{definition}

\begin{definition}[EFO1 query] The existential first-order logic query is defined as:
    \begin{align}
    \psi(y;x_1, \cdots, x_m) = \exists x_1, \cdots  x_m, \notag \\ 
    (c^1_1 \land \cdots \land c^1_{n_1}) \lor \cdots \lor (c^k_1 \land \cdots \land c^k_{n_k}),
\end{align}
where $c^i_{j}$ is atomic formula $r(h, t)$ or its negation $\neg r(h, t)$, $r \in\mathcal{R} $ is a relation predicate, $h$ and $t$ are entities belonging to $\mathcal{E}$ or a variable ranging from $\mathcal{E}$.
\end{definition}
\begin{definition}[Answer set]
    Given a query $\psi(y)$, its answer set is defined by 
    \begin{equation}
        \mathcal{A}[\psi(y)] = \{s \in \mathcal{E} | \psi |_{y=s} = \textsc{True}  \},
    \end{equation}
    where $\psi |_{y=s}$ represents substituting the occurrence of variable $y$ within $\psi$ to $s$  simultaneously.
\end{definition}
Under the Disjunctive Normal Form (DNF), an EFO1 query can be decomposed into a disjunction of conjunctive queries, where the answer set of the original query equals the union of the answer sets of the constituent conjunctive queries. Then we can mainly consider the conjunctive query $\phi$~\citep{ren_query2box_2020}. We can define the truth value function $T(\cdot)$ to infer formulas based on fuzzy logic~\citep{mendel1995fuzzy}.
\begin{definition}[Truth value function]\label{def:tv-formula}
    Let $\psi_1$ and $\psi_2$ be existential formulas, $\top_P$ and $\bot_P$ are product $t$-norms and $t$-conorms, and $r\in \mathcal{R}$, $s,o\in \mathcal{E}$, with $P_r(s, o)$ representing the truth value of atomic formula $r(s,o)$. The truth value function $T$, whose range is $[0, 1]$, is defined as follows:
    \begin{enumerate}[nosep]
    \item[(i)] $T(r(s, o))=P_r(s, o)$;
    \item[(ii)] $T(\lnot \psi_1) = 1 - T(\psi_1)$;
    \item[(iii)] $T(\psi_1 \land \psi_2) = T(\psi_1)~ \top_P~ T(\psi_2)$, where $a ~\top_{P}~ b = a \times b$;
    \item[(iv)] $T(\psi_1 \lor \psi_2) = T(\psi_1)~ \bot_P~ T(\psi_2)$, where a $~\bot_P~ b = 1-(1-a) \top_P (1-b)$;
    \item[(v)] $T(\exists x, \psi(x))=\max_{s\in \mathcal{E}} T(\psi|_{x=s})$.
    \end{enumerate}
\end{definition}
Then answering logical query can be rewritten as the following combinatorial optimization:
\begin{align}\label{eq:optimization}
T(\phi |_{y=s}) =  &\ max \quad T( \exists x_i, c_1|_{y=s} \land \cdots \land c_{n}|_{y=s}) \notag \\ 
\quad & \text{s.t.} \quad x_i \in \mathcal{E}, 1 \le i \le m.
\end{align}
We then introduce query graph to represent logical queries in a graph-structured form, allowing neural-symbolic methods to infer answers based on the graph topology.
\begin{definition}[Query graph]
The query graph of $\phi$ is defined as a set of quadruples, $G_{\phi} = \{(h_i, r_i, t_i, \textsc{Neg}_i)\}$. The $h$ and $t$ are nodes, where we refer to an entity as constant node and to variable as variable node. The edge is defined by a quadruple, along with two attributes: $r$, which denotes the relation, and $\textsc{Neg}_i$, which is the boolean variable indicating whether the atom is positive. 
\end{definition}

\subsection{Neural-symbolic search}
\begin{figure}[t]
    \centering
    \includegraphics[width=0.95\linewidth]{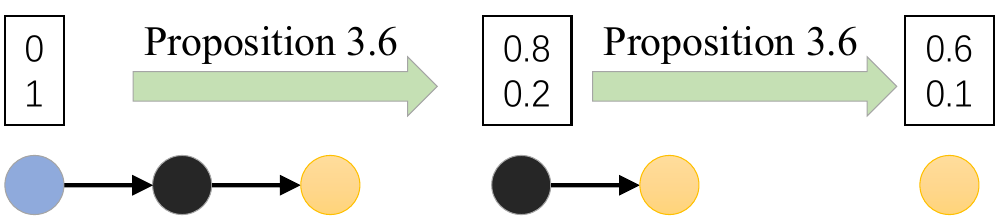}
    \caption{Execution steps of QTO/FIT for inferring a two-hop query. The fuzzy vectors of variables are iteratively updated as edges are removed. The final fuzzy vector of the free variable represents the predicted answer set.}\label{fig: FIT step pre}
\end{figure}

Existing neural-symbolic search methods, such as QTO~\citep{bai_answering_2023} and FIT~\citep{yin_rethinking_2023}, have shown that solving the combinatorial optimization in Equation~\ref{eq:optimization} is equivalent to performing edge-removal operations on the query graph. We provide an example in Figure~\ref{fig: FIT step pre} and a detailed proof can be found in Appendix~\ref{app:propagation details}. We introduce two important propositions to remove two kinds of nodes. 

\begin{proposition}[\textsc{RemoveConstNode}]\label{Remove_constant_nodes}
The constant node in $G_\phi$ can be removed in $\mathcal{O}(|\mathcal{E}|)$.

\end{proposition}
\begin{proposition}[\textsc{RemoveLeafNode}]\label{Remove_leaf_nodes}
 A leaf node is a variable node that connects to only one other variable node. The leaf node in $G_\phi$ can be removed in $\mathcal{O}(|\mathcal{E}|^2)$. 
\end{proposition}

If we index the entities in $\mathcal{E}=\{e_1,\ldots,e_{|\mathcal{E}|}\}$, we represent a \emph{fuzzy set over entities} for each variable $x$ using a \emph{fuzzy vector} $C_x \in [0,1]^{|\mathcal{E}|}$, where the $i$-th entry $C_x[i]$ denotes the membership degree of entity $e_i$ being a valid assignment for $x$ under current constraints. We define $\mu(s, C_x)$ to retrieve the membership degree of entity $s$ for variable~$x$.

Concretely, these methods initialize constant nodes with one-hot vectors and variable nodes with all-one vectors to record membership degrees. They then iteratively remove constant nodes along with their connected edges, while propagating the constraint information of the removed edges into the remaining nodes. Afterward, leaf nodes are successively removed. With this process, acyclic queries can be solved efficiently in $\mathcal{O}(n|\mathcal{E}|^2)$ time, where $n$ denotes the query size. However, for cyclic queries, FIT enumerates over the remaining variables, resulting in exponential complexity $\mathcal{O}(|\mathcal{E}|^n)$.

\section{Domain pruning via NLI} \label{Sec: neural logical index}
\begin{figure*}[t]
    \centering
    \includegraphics[width=0.95\linewidth]{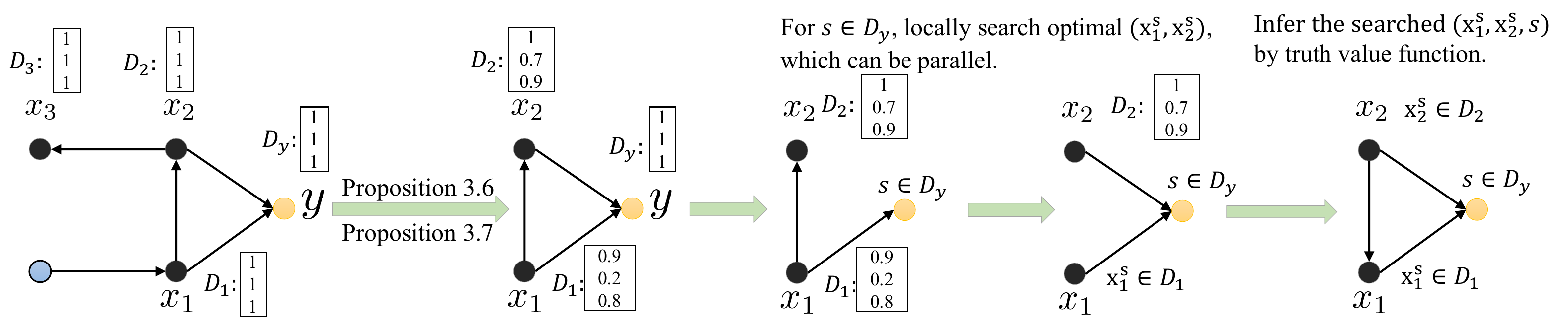}
    \caption{The execution steps of NLISA for inferring a cyclic query.}\label{fig:NLISA step pre}
\end{figure*}
\subsection{Motivation}
In this section, we introduce how to compute the pruned domain to obtain neural logical indices  for variables to accelerate symbolic search. The proposed neural logical indices have two main motivations. \textbf{(1) Only a small subset of entities is relevant to each variable.} As illustrated in Figure~\ref{fig:query graph and constraints}, for the logical query $\exists x_1,x_2, \text{Graduate}(x_1, x_2) \land  \text{Graduate}(x_1, y) \land  \text{Married}(x_2, y)$, the free variable $y$ should be consistent with both the ``Graduate'' and ``Married'' constraints. NLI therefore searches $y$ within a top-scored candidate set, such as $\{\text{Amy}, \text{Bob}, \text{Carol} \}$ in the toy example, instead of searching the entire space $\mathcal{E}$. This observation aligns with the classical idea of \emph{domain pruning} in constraint satisfaction problems (CSPs), where inconsistent values with respect to neighboring constraints are removed early~\citep{Chen2011ArcCA}. \textbf{(2) Knowledge graphs are typically sparse.} Due to the sparsity of connectivity in real-world KGs, the pruned domains can be significantly smaller than $\mathcal{E}$. As shown in Appendix~\ref{app:theory}, the empirical distribution of domain sizes follows a power law, meaning that most variables naturally correspond to small feasible domains. This property enables substantial reductions in symbolic search cost.


\subsection{Neural Logical Indices}





\begin{definition}[Neural Logical Indices (NLI)]
Given a variable $x$ in a query $\phi$, the NLI maps the subgraph $G^S_x$ centered at $x$ to a pruned domain $\mathcal{D}_x \subseteq \mathcal{E}$, representing the set of entities that remain plausible for $x$.
\end{definition}

In general, NLI first extracts a variable-centric subgraph $G^S_x$ from the original query graph $G_{\phi}$, determining the scope of constraints used. Then, we employ a neural embedding model to estimate how well each candidate entity satisfies the constraints encoded in $G^S_x$, producing neural scores over $\mathcal{E}$. Finally, we select the top-$k$ entities based on these scores to form the pruned domain $\mathcal{D}_x$.

NLI reduces the search space in Equation~\ref{eq:optimization}, providing the potential to accelerate complex symbolic search. This raises a central question: \textbf{\textit{how can we compute these reduced domains both efficiently and with a high compression ratio?}}

To  better utilize the trade-off between compression ratio and efficiency, we propose two strategies: \textbf{the local strategy prefers efficiency, while the global strategy emphasizes higher compression ratio.} An example in Figure~\ref{fig:query graph and constraints} illustrates the subgraph used by two constraint strategies.


\subsection{Local constraints strategy}
The local strategy restricts constraints to the 1-hop neighborhood subgraph defined below:
\begin{definition}[Neighbor subgraph]
    Let $G_{\phi}$ be the query graph. For any variable $x$ in $G_{\phi}$, we define the neighbor graph $G_n(x)$ as the subgraph consisting of all edges incident to $x$ and all nodes directly connected to $x$. The set of these adjacent nodes is denoted by $N_n(x)$.
\end{definition}

To further improve computational efficiency, we consider only relational information within the subgraph and discard node-specific features. We formulate a new task: predicting the tail set given a relation $r$ in the KG $\mathcal{G}$.
This task is analogous to link prediction in KGs. To enhance the generalization and training efficiency, we introduce a lightweight Hypernet~\citep{Ha2016HyperNetworks} that adapts existing KG embedding models~\citep{trouillon2016complex,chen2021relation} to this task. 
Given a relation--entity pair $(r,o)$, the likelihood of $o$ being a valid tail of $r$ is computed as $g_s(\texttt{HN}(r), \texttt{HN}(o))$, where $\texttt{HN}$ denotes the hyper-network and $g_s$ the adapted scoring function. 
We then apply fuzzy logic to combine relation likelihoods and infer the subgraph constraints. Further details of this task are provided in Appendix~\ref{app:details strategy}.


\subsection{Global constraints strategy}

The global constraint strategy considers all edges of the query graph while treating the target variable $x$ as the free variable. In other words, for the free variable $y$, we have $G^S_y = G_{\phi}$. Although this formulation increases the complexity of the subgraph, we can directly leverage the well studied query embedding methods to handle it. Further implementation details are provided in Appendix~\ref{app:details strategy}.  Following this, we apply a two-stage coarse-to-fine ranking procedure, similar to the approach used in information retrieval~\citep{Liu2019CoarsetoFineDR}.

\section{Efficient Search Method}\label{sec: effcient search}

We present the \textbf{Neural Logical Indices for Search Approximately (NLISA)} method, which integrates neural logical indices into neural-symbolic search to reduce computational cost. NLISA consists of two components: (i) domain pruning via NLI to accelerate node-removal operations, and (ii) an approximate local optimization procedure that replaces enumeration in cyclic structures. Figure~\ref{fig:NLISA step pre} illustrates the inference process on a representative cyclic query. We also provide the algorithm in Appendix~\ref{sec: algorithm of nlisa} to show the details of NLISA.

\subsection{Search with neural logical indices}

Given reduced domains $\{\mathcal{D}_x\}$, the removal of constant and leaf nodes can be performed over $\mathcal{D}_x$ rather than $\mathcal{E}$. The resulting complexities are
\[
\mathcal{O}(|\mathcal{D}_x|+|\mathcal{D}_y|), \quad
\mathcal{O}(|\mathcal{D}_x|^2+|\mathcal{D}_x||\mathcal{D}_y|),
\]
respectively. Implementations follow~\citep{bai_complex_2023,yin_rethinking_2023} with entities set replaced by their pruned domains (Appendix~\ref{app: fuzzy logic}). We empirically demonstrate that reduction factor $\kappa = \frac{|\mathcal{E}|}{|\mathcal{D}_x|}$ can be as large as $10$ in Section~\ref{sec: experimental results}, which leads to a significant acceleration in leaf-node removal. When $\kappa$ is fixed, the worst-case complexity for tree-form queries remains quadratic in $|\mathcal{E}|$; NLISA improves the constant factor by searching over $|\mathcal{E}|/\kappa$ candidates per variable. In practice, the effective $\kappa$ can be larger on sparse large KGs, so the observed search domain grows much more slowly than the entity set.

\subsection{Approximate search for cyclic queries}
\begin{table*}[t]
    \centering
    \caption{The statistics for the four knowledge graphs are provided, including the number of entities, relations, and edges. Additionally, we present the division of the training, validation, and test graphs.}\label{tab:KGs}
    \resizebox{0.8\linewidth}{!}{
    \begin{tabular}{lccccccc}
        \toprule
        Dataset & Entities & Relations & Training Edges & Validation Edges & Test Edges & Total Edges \\
        \midrule
        FB15K & 14,951 & 1,345 & 483,142 & 50,000 & 59,071 & 592,213 \\
        FB15K-237 & 14,505 & 237 & 272,115 & 17,526 & 20,438 & 310,079 \\
        NELL995 & 63,361 & 200 & 114,213 & 14,324 & 14,267 & 142,804 \\
        FB400K & 409,829 & 918 & 1,075,837 & 537,917 & 537,917 & 2,151,671 \\
        \bottomrule
    \end{tabular}}
\end{table*}
For cyclic queries, removing constant nodes and leaf nodes still leaves at least one cycle. Prior work resolves this by enumerating assignments over one remaining variable to break up the loop~\citep{yin_rethinking_2023}, resulting in exponential complexity. 

Instead of enumerating assignments to break cycles, we greedily assign variables in an order that maximizes local consistency, which empirically captures most of the global constraints while avoiding exponential search. For each candidate $s \in \mathcal{D}_y$, we optimize the assignment with highest likelihood conditioned on $y=s$:
\[
\mathop{\max}_{(x_1^s,\ldots,x_m^s)} \phi(s; x_1, \cdots, x_m).
\]
Without loss of generality, let the remaining variables be $\mathcal{X}_r=\{x_1,\ldots,x_m\}$, ordered by their shortest-path distances to the free variable $y$. This order is motivated by product t-norm semantics: multi-hop influence is aggregated by multiplying scores in $[0,1]$, so constraints closer to the free variable tend to have stronger influence on the final score. We therefore assign the nearest variables first and use later cycle-closing constraints mainly to filter locally consistent candidates. For each candidate $s\in\mathcal{D}_y$, we sequentially construct this assignment $(x_1^s,\ldots,x_m^s)$ using only local constraints. At step $i$, we have the partially updated query $\phi^{(i)}(s; x_1^s,\ldots,x_{i-1}^s,x_i, \cdots,x_m)$, and the assignment for $x_i$ is optimized as:
\[
x_i^s = \arg\max_{x_i}
    \left[
        \mathop{\top}_{e \in G_n(x_i, \phi^{(i)})} T(e)
    \right]
        \ \top\ 
        \mu(x_i, C_{x_i}),
\]
where $G_n$ denotes the neighborhood subgraph of $x_i$, $T(\cdot)$ is the truth-value function,  and $\mu(\cdot,C_{x_i})$ records the membership degrees of each candidate value. The neighborhood subgraph contains the essential constraints for the searched variable, and restricting to its first-order connections prevents complexity from propagating to higher-order neighbors. Under this restriction, the per-step computational complexity is $\mathcal{O}\left(|\mathcal{D}_{x_i}|^2+|\mathcal{D}_{x_i}||\mathcal{D}_y|\right)$.

After $m$ steps we obtain $\phi^{(m+1)}$, and compute $\phi\mid_{y=s}$ over the approximate searched assignments $(x_1^s,\ldots,x_m^s)$ via t-norm aggregation. Candidates $s\in\mathcal{D}_y$ are processed independently and can be evaluated in parallel. This local procedure does not provide a worst-case optimality guarantee, but it is effective for the short cyclic structures in current CQA benchmarks; more complex or longer cycles may require larger domains or stronger search heuristics. We denote this procedure as \textsc{LocalOptimize} (see Figure~\ref{fig:NLISA step pre}) and provide its algorithm in Appendix~\ref{sec: algorithm of nlisa}.

\begin{algorithm}[t]
\caption{NLISA: Neural Logical Indices for Approximate Search}\label{alg: nlisa}
\label{alg:nlisa}
\begin{algorithmic}[1]
\REQUIRE Conjunctive query $\phi$ with free variable $y$, pretrained KG embeddings, reduction factor $\kappa$
\ENSURE Fuzzy vector of free variable $C_y \in \mathcal{D}_y^{[0,1]}$ 

\STATE \textbf{Neural logical indexing}
\FOR{each variable $x$ in $\phi$}
    \STATE Compute fuzzy scores $S_x$ from the subgraph $G_x^S$
    \STATE $\mathcal{D}_x \leftarrow \text{Top-}\lfloor \frac{|\mathcal{E}|}{ \kappa} \rfloor(S_x)$
\ENDFOR
\STATE Initialize fuzzy vectors $\{C_x\}$ and $C_y$ based on $\mathcal{D}_x$ and $\mathcal{D}_y$

\STATE \textbf{Query simplification}
\WHILE{$\phi$ contains constant or leaf variables}
    \STATE Remove the node and propagate constraints into $\{C_x\}$
    \STATE $\phi$ is updated.

\ENDWHILE

\STATE \textbf{Cyclic query handling}
\IF{$\phi$ is acyclic}
    \STATE Use the fuzzy vector $C_y$ induced by $\phi$ as the output
\ELSE
    \STATE $\{(s;x_1^s,\ldots,x_m^s)\} \leftarrow \textsc{LocalOptimize}(\phi,\{\mathcal{D}_x\},\{C_x\},y)$
    \STATE Update the fuzzy vector $C_y$ on $\{\phi(s;x_1^s,\ldots,x_m^s)\}$  based on the searched assignments $\{(s;x_1^s,\ldots,x_m^s)\}$
\ENDIF

\RETURN Fuzzy vector $C_y$ as the predicted answer set.
\end{algorithmic}
\end{algorithm}

\subsection{Algorithm and complexity}

We summarize the overall procedure of NLISA for inferring EFO1 queries in Algorithm~\ref{alg: nlisa}. We then analyze the space and time complexity of the proposed method.

The space complexity of our method is $\mathcal{O}((|\mathcal{E}| + |\mathcal{R}| +d_h) \times d) $, where $d$ is the embedding dimension of the KGE models and $d_h$ is the hidden dimension of hyper-network. This complexity is on the same order as standard KGE models, which scales linearly with the sizes of entities and relations.

Inferring queries with $n$ atom formulas, the time complexity is $\mathcal{O}((|\mathcal{D}_x| |\mathcal{D}_y|  + |\mathcal{D}_x|^2)nd)$, where $|\mathcal{D}_x|$ and $|\mathcal{D}_y|$ are the size of search domain of the existential and free variables, respectively, and dominated lower-order terms are omitted. Our algorithm \textbf{exhibits quadratic complexity with respect to the pruned domain size for EFO1 queries}.

\begin{table*}[t]
\centering
\caption{The results of efficiency (QPS) and effectiveness (MRR) on the tree-form queries under $\kappa \approx 10$. We employ BetaE benchmark~\citep{ren_beta_2020} for FB15K-237, FB15K, and NELL, as well as utilize Smore benchmark~\citep{ren_smore_2022} for FB400K. Best results in each row are shown in bold.}
\label{tab: overall results over kgs}
\resizebox{0.7\linewidth}{!}{
\begin{tabular}{cccccccc}
\toprule
KG                     & Entities &     Metric            & BetaE  & CQD-CO  & QTO/FIT   & NLISA (L) & NLISA (G) \\ \midrule
\multirow{2}{*}{FB15K-237} & \multirow{2}{*}{14,505} & QPS~$\uparrow$   & 100     & 14   & 31  & \textbf{115}   & 86   \\
   & & MRR~$\uparrow$ & 16.1    & 15.2  & \textbf{23.1}  & 21.2  & 22.8  \\ \midrule
\multirow{2}{*}{FB15K} & \multirow{2}{*}{14,951}    & QPS~$\uparrow$  &    97        &  10.1    &  20     &          \textbf{109}  &     79        \\
                        &   & MRR~$\uparrow$ &  32.4       &  32.5    &  61.1     &   56.4         &   \textbf{61.9}  \\ \midrule
\multirow{2}{*}{NELL}  & \multirow{2}{*}{63,361}    & QPS~$\uparrow$  & 23       & 7    & 3  & \textbf{35}  & 20\\

                        &   & MRR~$\uparrow$ & 18.8   & 19.9 & 27.2 & 27.0 & \textbf{27.3}       \\ \bottomrule        
\multirow{2}{*}{FB400K}   & \multirow{2}{*}{409,829}   & QPS~$\uparrow$  & {5.4}  & 1.6    & OOM & \textbf{8.2}  &\\

                        &   & MRR~$\uparrow$ & 50.5  & 43.3 & OOM & \textbf{58.9} &        \\ \bottomrule 
\end{tabular}}
 \vspace{-1em}

\end{table*}

\section{Experimental settings}
\subsection{Datasets}
Standard CQA datasets consist of queries sampled from knowledge graphs (KGs) according to predefined abstract query types. Table~\ref{tab:KGs} reports the statistics of the KGs used in our experiments, including the numbers of entities, relations, and edges in each split. In CQA, a query type is an abstract logical template that characterizes the structural form of a query, including its logical operators and variable dependencies, independent of specific entities or relations. Different query types capture distinct reasoning patterns and therefore pose fundamentally different computational challenges.

The BetaE benchmark~\citep{ren_beta_2020} is a widely used CQA benchmark that primarily focuses on tree-form queries. It is constructed over three KGs: FB15K~\citep{bordes_translating_2013}, FB15K-237~\citep{toutanova_representing_2015}, and NELL995~\citep{xiong2022faithful}. In total, the BetaE benchmark includes 14 distinct query types, among which 5 involve negation operations.

The real EFO1 benchmark~\citep{yin_rethinking_2023} extends BetaE by introducing 10 additional query types beyond tree-form queries, while using the same underlying KGs. In particular, real EFO1 incorporates more complex logical structures, including multi-branch and cyclic query graphs. Visualizations of the query types in both BetaE and real EFO1 are provided in Appendix~\ref{app:details query types}.

The Smore benchmark~\citep{ren_smore_2022} considers the same tree-form query types as BetaE but samples queries from a much larger-scale KG. Since only the FB400K dataset with approximately 400{,}000 entities has been released, we adopt FB400K as our large-scale benchmark.

\subsection{Evaluation protocol}
Following the standard evaluation protocol for CQA over incomplete KGs~\citep{ren_beta_2020}, we evaluate models using ranking-based metrics by distinguishing query answers into \emph{easy} and \emph{hard} sets. Hard answers correspond to non-trivial cases that require predicting at least one missing link not observed in the training or validation graphs. We report Mean Reciprocal Rank (MRR) and Hits@K~\citep{hamilton_embedding_2018}.

Given an observed KG $\mathcal{G}$ with predefined splits $\mathcal{G}_{\text{train}} \subset \mathcal{G}_{\text{valid}} \subset \mathcal{G}_{\text{test}}$,
all splits share the same entity set $\mathcal{E}$ and relation set $\mathcal{R}$\citep{ren_beta_2020}. For a query $q$, let $[q]_{\text{train}}$, $[q]_{\text{valid}}$, and $[q]_{\text{test}}$ denote its answer sets on the corresponding graphs. Hard answers are defined as
$[q]_{\text{hard}} = [q]_{\text{test}} \setminus [q]_{\text{valid}}.$
Each hard answer is ranked against candidate entities in $\mathcal{E} \setminus [q]_{\text{valid}}$. Given the resulting rank $r$, we compute MRR as $1/r$ and Hits@K as $\mathbb{I}(r \le K)$.

To evaluate efficiency, we report the averaged Queries Per Second (QPS). To ensure fair comparison, we randomly select the 100 queries for each query type, which reduces the bias caused by distributions of query types.

\subsection{Baselines}
We consider representative and strong CQA models as baselines from four categories: query embedding methods (BetaE~\citep{ren_beta_2020} and ConE~\citep{zhang_cone_2021}), graph neural network method (LMPNN~\citep{wang_logical_2023} and GNN-QE~\citep{zhu_neural-symbolic_2022}), and symbolic search method (CQD~\citep{arakelyan_complex_2020}, QTO~\citep{bai_answering_2023}, and FIT~\citep{yin_rethinking_2023}). To ensure fairness, we use the same checkpoint for CQA, QTO, FIT, and our method. Since FIT is equivalent to QTO in tree-form queries~\citep{yin_rethinking_2023}, we do not distinguish them in this case. For a fair comparison, we exclude LLM-based methods from our baselines, as they introduce additional parametric knowledge beyond the observed KG. Moreover, existing LLM-based approaches to CQA typically support only DAG-structured queries relying on tree decomposition and are not designed to handle general $\text{EFO}_1$ queries.


\subsection{Implementation details}
Both NLISA (L) and NLISA (G) utilize the local constraints strategy for existential variables. NLISA (L) applies the local strategy for free variables, while NLISA (G) solves the global constraints by LMPNN~\citep{wang_logical_2023}. Thus, NLISA (G) is the accuracy-oriented variant, whereas NLISA (L) is the faster and more scalable variant. On FB400K, we report NLISA (L) because training the LMPNN component used by NLISA (G) requires full-graph message passing over the entity set and exceeds GPU memory. For more details of the implementation, see Appendix~\ref{app: implementation}.

\section{Experimental results}~\label{sec: experimental results}
We show that NLISA achieves substantial speedup with modest performance loss by reducing the search domain (large $\kappa$). First, we present the overall results across various KGs with $\kappa \approx 10$, highlighting NLISA's scalability to the larger FB400K dataset.  Second, we provide MRR and efficiency results across various query types, particularly illustrating that NLISA significantly accelerates cyclic queries. Lastly, we conduct an ablation study of $\kappa$ under different settings to further analyze the effectiveness of domain pruning. 

Appendix~\ref{app:additional result} provides extended experiments on efficiency and memory usage, domain-size sensitivity, KGE backbones, recent query embedding baselines, and a hard-query benchmark.

\begin{table*}[t]
\centering
\scriptsize
\caption{MRR results (\%) for various query types under $\kappa \approx 10$. We include tree-form queries without negation from the BetaE benchmark~\citep{ren_beta_2020}, where $\text{A}_{\text{P}}$ denotes the average for these query types. Additionally, we incorporate all queries from the real EFO1 benchmark~\citep{yin_rethinking_2023}, with $\text{A}_{\text{R}}$ representing the average for these query types. The best results are highlighted in red, while the second-best results are shown in blue.}
\label{tab:tree form (p) and real efo1}
\resizebox{\linewidth}{!}{
\input{tabs/tab1}
}
\end{table*}

\begin{table}[ht]
\centering
\caption{MRR results (\%) of negation queries from tree-form BetaE benchmark under $\kappa \approx 10$, where $\text{A}_{\text{N}}$ denotes the average for these query types. We highlight the best results in red and the second-best results in blue. }
\label{tab:tree form (n)}
\resizebox{\linewidth}{!}{
\input{tabs/tab2}
}
\end{table}

\subsection{Overall results on various KGs}
\paragraph{Setup.} We first demonstrate that the domain size in symbolic search can be substantially reduced without sacrificing the accuracy of inferring unseen answers. Accordingly, we restrict the candidate domain to less than 10\% of all entities and set $|\mathcal{D}_x| = |\mathcal{D}_y|$ to 2000, 2000, 6000, and 8000 for FB15K-237, FB15K, NELL, and FB400K, respectively. The corresponding results are reported in Table~\ref{tab: overall results over kgs}.

\paragraph{Analysis.} From a methodological perspective, our NLISA consistently yields strong results in both QPS and MRR compared to baseline methods. The efficient query embedding method (BetaE) exhibits low MRR, while existing symbolic search methods (QTO and FIT) are limited in QPS. NLISA (L) demonstrates the best efficiency, whereas NLISA (G) typically achieves the highest MRR. Regarding KGs, larger entity sets pose greater efficiency challenges for baseline models. Due to their quadratic complexity, QTO and FIT exhibit low QPS for larger graphs and even face OOM issues with FB400K. Across FB15K, NELL, and FB400K, the selected domain sizes are 2K, 6K, and 8K, respectively, while the entity set grows from about 15K to 400K. This indicates that the effective search domain grows much more slowly than the entity set on these sparse KGs. For embedding methods, training query embedding methods is particularly challenging. Notably, NLISA (L) relies solely on KGE models and can be easily deployed for various sizes of KGs.

\subsection{Performance on various query types}\label{Sec: results betae}

\paragraph{Setup.} Using the same experimental setup, we further report the performance of NLISA across different query types to analyze its behavior under varying query structures\footnote{The query type ``pni'' corresponds to a universal first-order logic query and is therefore excluded from evaluation.}. The results for positive queries in the BetaE benchmark and all queries in the real EFO1 benchmark are reported in Table~\ref{tab:tree form (p) and real efo1}, while the results for negative queries are presented in Table~\ref{tab:tree form (n)}.

\paragraph{Analysis.} By reducing the search domain with $\kappa \approx 10$ and employing fast approximate search, our method achieves remarkable efficiency without compromising MRR across a wide spectrum of query types.  Specifically, compared to the previous SoTA method FIT, NLISA (G) achieves 99\% relative MRR on positive tree-form queries and 95\% relative MRR on general EFO1 queries. This shows that the potential performance drop of NLISA is small. In particular, for negation queries and complex EFO1 types like ``2il'' and ``3il'', NLISA (G) even outperforms FIT, which demonstrates that reduced search domains can even improve the accuracy by effectively filtering out noise. Negative queries have a high noise-to-signal ratio because many entities can trivially satisfy a negated condition. NLI first recalls the top-scored candidates with the embedding model and removes many low-relevance candidates, after which symbolic search performs ranking on a cleaner domain. Regarding cyclic queries, including ``3c'' and ``3cm'', our approximate search method achieves an average performance of nearly 95\% compared to FIT, validating its effectiveness. The remaining gap is expected because FIT performs exact enumeration at exponential cost, whereas NLISA uses quadratic-time approximate local search.

Although NLISA (L) produces slightly lower performance compared to NLISA (G), it offers significantly higher QPS and greater convenience, making it practical for real-world applications.

\subsection{Efficiency results on various query types}\label{Sec: results efo1}

\begin{figure*}[t]
    \centering
    \begin{minipage}{0.48\textwidth}
        \centering
        \includegraphics[width=\linewidth]{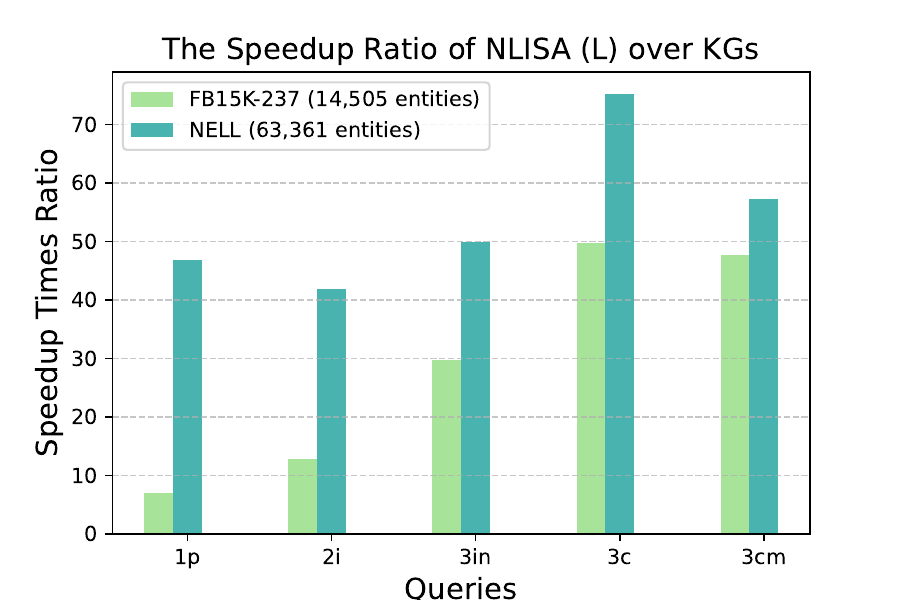}
    \end{minipage}\hfill
    \begin{minipage}{0.48\textwidth}
        \centering
        \includegraphics[width=\linewidth]{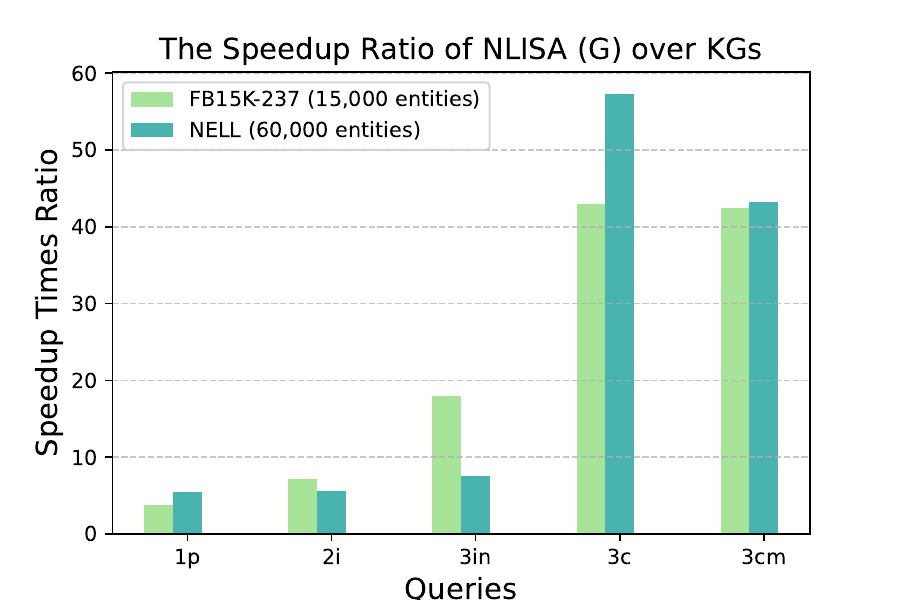}
    \end{minipage}
    \caption{The efficiency results of NLISA for various query types on FB15K-237 and NELL. The included query types are 1p, 2i, 3in, 3c, and 3cm, arranged according to their query complexity. Notably, 3c and 3cm are cyclic queries with NP-hard complexity.}
    \label{fig: efficiency query}
\end{figure*}
To analyze the speedup performance across different query types, we present the speedup ratios compared to FIT in Figure~\ref{fig: efficiency query}. We selected five representative query types, arranged by the complexity of symbolic search. We conduct the ablation study over two methods (NLISA (L), NLISA (G)) and two KGs (FB15K-237, and NELL). 

The results indicate that cyclic queries, such as 3c and 3cm, exhibit significantly larger speedup ratios compared to other tree-form queries across various KGs and methods, validating the effectiveness of our approximate search algorithm. Notably, the speedup ratio of NLISA (L) surpasses that of NLISA (G), especially in the NELL dataset with its larger entity set, highlighting the effectiveness of the local constraints strategy.

\subsection{Study on pruned domain sizes and $\kappa$}

We conduct the ablation study on $\kappa$ in NLISA to investigate the empirical impact of varying domain sizes.
We report results across different KGs in Table~\ref{tab:ab domain on kg}, and present detailed results of NLISA (L) and NLISA (G) on NELL in Table~\ref{tab: ab domain on local and global}. Both MRR and running time are reported, with additional details provided in the Appendix~\ref{app:additional result}.

The results in Table~\ref{tab: ab domain on local and global} indicate that both the running time and MRR of NLISA (L) and NLISA (G) increase with the growth of domain size. At a domain size of 2000, which represents 3\% of the entity set, NLISA (G) achieves an MRR of 99\%, while NLISA (L) reaches 95\%. This validates the sparse nature of knowledge graphs and underscores the effectiveness of our domain reduction strategy. Furthermore, Table~\ref{tab:ab domain on kg} indicates that the marginal gains in MRR diminish as the domain size continues to increase. This observation suggests that $\kappa$ can be selected using validation performance in practice, or determined in a more adaptive manner based on the score distribution produced across $\kappa$. Smaller domains bring higher speed, while larger domains reduce the risk of pruning correct answers, especially when the KG is extremely sparse or the underlying embeddings have lower recall.

\begin{table}[t]
    \centering
    \caption{Ablation study on different $\kappa$ size on NLISA (G) and NLISA (L) under the NELL dataset. We report the total running times (in seconds) and MRR results.}\label{tab: ab domain on local and global}
    \resizebox{0.98\linewidth}{!}{
    \begin{tabular}{cccccccc}
        \toprule
       Method & Dataset & 1,000 & 2,000 & 4,000  & 6,000  & 8,000 &  \\
        \midrule
        \multirow{2}{*}{NLISA (L)}  & Time& 509 & 640 & 1008 & 1429 & 2351 &  \\
        & MRR & 30.3 & 32.3 & 33.6 & 33.7 & 33.8 &  \\ \midrule
        \multirow{2}{*}{NLISA (G)} & Time  & 2821 & 2823 & 2856 & 2856 & 3944 &  \\
        & MRR & 33.6 & 33.8 & 33.9 & 33.9 & 33.9 &  \\
        \bottomrule
    \end{tabular}}
\vspace{-2em}

\end{table}

\begin{table}[t]
\centering
\caption{Ablation study on different $\kappa$ size on NLISA (G) under different KGs. We report the total running times (in seconds) and MRR results.}\label{tab:ab domain on kg}
\resizebox{0.98\linewidth}{!}{
\begin{tabular}{l l c c c c c}
\toprule
\textbf{Dataset} & \textbf{Metric}
& \textbf{1000} & \textbf{2000} & \textbf{4000} & \textbf{6000} & \textbf{8000} \\
\midrule
\multirow{3}{*}{FB15K-237}
& $\kappa$ & 14.5 & 7.3 & 3.6 & 2.4 & 1.8 \\
& Times    & 382.5  & 454.2  & 1390.9  & 2024.8 & 3305.4  \\
& MRR      & 23.8   & 25.4   & 26.4    & 26.6   & 26.6   \\
\midrule
\multirow{3}{*}{FB15K}
& $\kappa$ & 15.0 & 7.5 & 3.7 & 2.5 & 1.9 \\
& Times    & 2232.4 & 3197.2 & 4177.5  & 4188.2 & 6720.2  \\
& MRR      & 59.2   & 71.1   & 71.3    & 71.9   & 72.1   \\
\midrule
\multirow{3}{*}{NELL}
& $\kappa$ & 63.4 & 31.7 & 15.8 & 10.6 & 7.9 \\
& Times    & 2820.8 & 2822.9  & 2856.6   & 3263.2 & 3944.2  \\
& MRR      & 33.6   & 33.8    & 33.9     & 33.9   & 33.9   \\
\bottomrule
\end{tabular}
}
\vspace{-2em}
\end{table}

\section{Limitations and future work}
NLISA is designed as an accuracy--efficiency trade-off for neural-symbolic CQA rather than an exact replacement for exhaustive search. For tree-form queries, when $\kappa$ is treated as a constant, the worst-case complexity remains quadratic in the entity-set size; the practical gain comes from the smaller effective search domain. For cyclic queries, local search avoids exponential enumeration but is approximate and does not provide a worst-case optimality guarantee. Current benchmarks mainly contain short cycles, where NLISA retains most of FIT's MRR; longer cycles or multiple overlapping cycles may lead to larger approximation gaps and are important future directions. NLISA also depends on the quality of the underlying KGE or query embedding scores used for pruning. Poor embeddings may prune correct answers, although this can be mitigated by using a larger search domain. Finally, NLISA (G) is limited on very large KGs because its LMPNN component requires full-graph message passing; scaling this variant and evaluating on larger and more diverse KGs such as Wikidata are left for future work.

\section{Conclusion}
Although symbolic search methods exhibit strong performance in CQA tasks, they face efficiency challenges that hinder further development and application. Query embedding models offer fast speed but often provide generic performance. In this paper, we integrate their respective advantages and propose an approximate search method combined with neural logical indices. The neural logical indices can be computed rapidly using embedding methods, significantly reducing the effective search domain of symbolic methods. In particular, the approximate search method can handle cyclic queries with quadratic complexity in the pruned domains. Experiments on various benchmarks show that, with a 10$\%$ reduced search domain, our method preserves most of the MRR of neural-symbolic search while substantially improving QPS. Additionally, we demonstrate that our method can execute on a KG with an order of magnitude more entities than before, highlighting its practical scalability.

%% file: tabs/tab1.tex
\begin{tabular}{cccccccccccccccccccccc}
\toprule
  Method & 1p & 2p & 3p & 2i & 3i & ip & pi & 2u & up & $\text{A}_{\text{P}}$ & pni & 2il & 3il & 2m & 2nm & 3mp & 3pm & im & 3c & 3cm & $\text{A}_{\text{R}}$ \\ \midrule    
\multicolumn{22}{c}{FB15K-237} \\ \midrule
 CQD-CO & \textcolor{red}{\textbf{46.7}} & 9.6 & 6.2 & 31.2 & 40.6 & 16.0 & 23.6 & 14.5 & 8.2 & 21.9 & 7.7 & 29.6 & 46.1 & 6.0 & 1.7 & 6.8 & 3.3 & 12.3 & 25.9 & 23.8 & 16.3\\
 ConE& 41.8 & 12.8 & 11.0 & 32.6 & 47.3 & 25.5 & 14.0 & 14.5 & 10.8 & 23.4 & 10.8 & 27.6 & 43.9 & 9.6 & 7.0 & 9.3 & 7.3 & 14.0 & 28.2 & 24.9 & 18.3\\
  LMPNN & 45.9 & 13.1 & 10.3 & 34.8 & 48.9 & 17.6 & 22.7 & 13.5 & 10.3 & 24.1 & 10.7 & 28.7 & 42.1 & 9.4 & 4.2 & 9.8 & 7.2 & 15.4 & 25.3 & 22.2 & 17.5 \\ \midrule[0.2pt] 
   GNN-QE & 42.8 & \textcolor{red}{\textbf{14.7}} & 11.8 & \textcolor{red}{\textbf{38.3}} & \textcolor{red}{\textbf{54.1}} & 18.9 & \textcolor{red}{\textbf{31.1}} & 16.2 & \textcolor{red}{\textbf{13.4}} & 26.8 \\

  QTO & \textcolor{red}{\textbf{46.7}} & \textcolor{blue}{\textbf{14.6}} & \textcolor{red}{\textbf{12.8}} & \textcolor{blue}{\textbf{37.5}} & \textcolor{blue}{\textbf{51.6}} & \textcolor{red}{\textbf{21.9}} & 30.1 & \textcolor{red}{\textbf{18.0}} & \textcolor{blue}{\textbf{13.1}} & \textcolor{red}{\textbf{27.4}} & 12.1 & 28.9 & 47.9 & 8.5 & 10.7 & 11.4 & 6.5 & 17.9 & 38.3 & 35.4 & 21.8 \\

  FIT & \textcolor{red}{\textbf{46.7}} & \textcolor{blue}{\textbf{14.6}} & \textcolor{red}{\textbf{12.8}} & \textcolor{blue}{\textbf{37.5}} & \textcolor{blue}{\textbf{51.6}} & \textcolor{red}{\textbf{21.9}} & 30.1 & \textcolor{red}{\textbf{18.0}} & \textcolor{blue}{\textbf{13.1}} & \textcolor{red}{\textbf{27.4}} & \textcolor{red}{\textbf{14.9}} & \textcolor{blue}{\textbf{34.2}} & \textcolor{blue}{\textbf{51.4}} & \textcolor{red}{\textbf{9.9}} & \textcolor{red}{\textbf{12.7}} & \textcolor{red}{\textbf{11.9}} & \textcolor{red}{\textbf{7.7}} & \textcolor{red}{\textbf{19.6}} & \textcolor{red}{\textbf{39.4}} & \textcolor{red}{\textbf{37.3}} & \textcolor{red}{\textbf{23.9}}\\

  NLISA (L) &46.6 & 13.8 & 12.2 & 33.1 & 45.8 & 20.2 & 27.1 & 16.9 & 11.4 & 25.2& 12.5 & 34.0 & 51.8 & 9.2 & \textcolor{blue}{\textbf{9.8}} & 10.3 & 7.4  & 17.8 & 34.9 & 34.5 & 22.2  \\
  NLISA (G) &  46.6 & 14.1 & 12.3 & 37.2 & 51.3 & \textcolor{blue}{\textbf{21.3}} & \textcolor{blue}{\textbf{30.2}} & \textcolor{blue}{\textbf{17.6}} & 12.2 & \textcolor{blue}{\textbf{27.0}}&\textcolor{blue}{\textbf{14.2}} & \textcolor{red}{\textbf{34.3}} & \textcolor{red}{\textbf{52.7}} & \textcolor{blue}{\textbf{9.5}} & 9.2  & \textcolor{blue}{\textbf{10.7}} & \textcolor{blue}{\textbf{7.4}} & \textcolor{blue}{\textbf{18.1}} & \textcolor{blue}{\textbf{36.0}} & \textcolor{blue}{\textbf{35.5}} & \textcolor{blue}{\textbf{22.8}} 

 \\ \midrule
 \multicolumn{22}{c}{FB15K} \\ \midrule
  CQD-CO & 89.2& 25.6 & 13.6 & 77.4 & 78.3 & 44.2 & 33.2 & 41.7 & {22.1} & 46.9 & 24.2 & 47.6 & 65.4 & 23.2 & 1.6 & 11.0 & 8.7 & 36.3 & 31.3 & 32.9 & 28.2 \\
 ConE& 75.3 & 33.8 & 29.2 & 64.4 & 73.7 & 50.9 & 35.7 & 55.7 & 31.4 &49.8 & 37.0 & 40.1 & 57.3 & 33.3 & 11.5 & 23.9 & 27.6 & 38.7 & 35.0 & 36.3 & 34.1  \\
  LMPNN & 85.0 & 39.3 & 28.6 & 68.2 & 76.5 & 43.0 & 46.7 & 36.7 & 31.4 & 50.6 & 38.7 & 43.2 & 57.8 & 40.3 & 7.9 & 24.0 & 30.5 & 48.4 & 32.2 & 30.9 & 35.4\\  \midrule[0.2pt] 
  GNN-QE & 88.5 & \textcolor{red}{\textbf{69.3}} & \textcolor{red}{\textbf{58.7}} & \textcolor{red}{\textbf{79.7}} & \textcolor{red}{\textbf{83.5}} & 70.4 & 69.9 & \textcolor{red}{\textbf{74.1}} &\textcolor{red}{ \textbf{61.0}} & \textcolor{red}{\textbf{72.8}}  \\ 
   QTO &  \textcolor{red}{\textbf{89.4}} & \textcolor{blue}{\textbf{65.6}} &\textcolor{blue}{ \textbf{56.9}} & \textcolor{blue}{\textbf{79.1}} & \textcolor{red}{\textbf{83.5}} & \textcolor{red}{\textbf{71.8}} & \textcolor{red}{\textbf{73.1}} & \textcolor{blue}{\textbf{73.9}} & 59.0 & \textcolor{blue}{\textbf{72.5}}  & 48.2 & 49.5 & 68.2 & 64.6 & 19.4 & 48.5 & 53.7 & 73.9 & 53.3 & 54.9 & 53.4  \\

   FIT &  \textcolor{red}{\textbf{89.4}} & \textcolor{blue}{\textbf{65.6}} &\textcolor{blue}{ \textbf{56.9}} & \textcolor{blue}{\textbf{79.1}} & \textcolor{red}{\textbf{83.5}} & \textcolor{red}{\textbf{71.8}} & \textcolor{red}{\textbf{73.1}} & \textcolor{blue}{\textbf{73.9}} & 59.0 & \textcolor{blue}{\textbf{72.5}} & \textcolor{blue}{\textbf{57.9}} & \textcolor{red}{\textbf{70.4}} & \textcolor{blue}{\textbf{77.6}} & \textcolor{red}{\textbf{73.5}} & \textcolor{red}{\textbf{39.1}} & \textcolor{red}{\textbf{57.3}} & \textcolor{red}{\textbf{64.0}} &\textcolor{red}{ \textbf{79.4}} & \textcolor{red}{\textbf{63.8}} & \textcolor{red}{\textbf{65.4}} & \textcolor{red}{\textbf{64.8}}  \\

    NLISA (L) &  87.2 & 61 & 53.1 & 67.7 & 75.4 & 66.2 & 65 & 56.3 & 46.7 & 64.3 & {55.9} & {69.1} & {76.0} & {66.0} & \textcolor{blue}{\textbf{38.3}} & {49.1} & {55.2} & {70.5} & {57.8} & {59.7} & {59.8} \\ 

  NLISA (G) & \textcolor{red}{\textbf{89.4}} & 64.7 & 55.4 & 76.4 & 82.3 & 71.1 & \textcolor{blue}{\textbf{71.1}} & 71.9 & 57.5 & 71.1 &  \textcolor{red}{\textbf{60.7}} & \textcolor{red}{\textbf{70.4}} & \textcolor{red}{\textbf{78.0}} & \textcolor{blue}{\textbf{68.8}} & {36.2} & \textcolor{blue}{\textbf{51.0}} & \textcolor{blue}{\textbf{57.0}} & \textcolor{blue}{\textbf{76.1}} & \textcolor{blue}{\textbf{60.3}} & \textcolor{blue}{\textbf{61.5}} & \textcolor{blue}{\textbf{62.0}}  \\

     \midrule
  \multicolumn{22}{c}{NELL} \\ \midrule
  CQD-CO & 60.4 & 17.8 & 12.8 & 39.3 & 46.6 & 22.1 & 30.1 & 17.3 & 13.2 & 28.8 & 7.9 & 48.7 & 68.0 & 31.7 & 1.5 & 12.9 & 13.8 & 33.9 & 38.8 & 35.9 & 29.3 \\
 ConE & 53.1 & 16.1 & 13.9 & 40.0 & 50.8 & 26.3& 17.5 & 15.3 & 11.3  & 27.2 & 10.3 & 42.1 & 65.8 & 32.4 & 7.0 & 12.6 & 16.8 & 34.4 & 40.2 & 38.2 & 30.0 \\
  LMPNN & 60.6 & 22.1 & 17.5 & 40.1 & 50.3 & 24.9 & 28.4 & 17.2 & 15.7 & 30.8& 11.6 & 43.9 & 62.3 & 35.6 & 6.2 & 15.9 & 19.3 & 38.3 & 39.1 & 34.4 & 30.7 \\ \midrule[0.2pt] 
   GNN-QE & 53.3 & 18.9 & 14.9 & 42.4 & 52.5 & 18.9 & 30.8 & 15.9 & 12.6 & 28.9 \\ 
  QTO & \textcolor{red}{\textbf{60.8}} & \textcolor{red}{\textbf{23.8}} & \textcolor{red}{\textbf{21.2}} & \textcolor{blue}{\textbf{44.3}} & \textcolor{blue}{\textbf{54.1}} & 26.6 & {31.7} & \textcolor{red}{\textbf{20.3}} & \textcolor{red}{\textbf{17.6}} & \textcolor{blue}{\textbf{33.4}} & 12.3 & 48.5 & 68.2 & 38.8 & 12.3 & 22.8 & 19.3 & 41.1 & 45.4 & 43.9 & 35.3\\

    FIT & \textcolor{red}{\textbf{60.8}} & \textcolor{red}{\textbf{23.8}} & \textcolor{red}{\textbf{21.2}} & \textcolor{blue}{\textbf{44.3}} & \textcolor{blue}{\textbf{54.1}} & 26.6 & {31.7} & \textcolor{red}{\textbf{20.3}} & \textcolor{red}{\textbf{17.6}} & \textcolor{blue}{\textbf{33.4}} & \textcolor{red}{\textbf{14.4}} & \textcolor{blue}{\textbf{53.3}} & {69.5} & \textcolor{red}{\textbf{42.1}} & \textcolor{red}{\textbf{12.5}} & \textcolor{red}{\textbf{24.0}} & {22.8} & \textcolor{red}{\textbf{41.5}} & \textcolor{red}{\textbf{47.5}} &\textcolor{red}{ \textbf{45.3}} & \textcolor{red}{\textbf{37.3}} \\

   NLISA (L) &     60.4 & 23.5 & \textcolor{blue}{\textbf{20.1}} & 43.7 & \textcolor{blue}{\textbf{54.1}} & \textcolor{blue}{\textbf{27.0}} & \textcolor{blue}{\textbf{32.8}} & \textcolor{red}{\textbf{20.3}} & 17.3 & 33.2 & \textcolor{blue}{\textbf{14.0}} & 53.1 & \textcolor{blue}{\textbf{71.7}}  & 23.0 & \textcolor{blue}{\textbf{11.8}} & \textcolor{blue}{\textbf{23.0}} & \textcolor{blue}{\textbf{23.0}} & 41.1 & \textcolor{blue}{\textbf{45.6}} & \textcolor{blue}{\textbf{44.4}} & \textcolor{blue}{\textbf{36.8}}  \\ 

 NLISA (G) &  \textcolor{red}{\textbf{60.8}} & \textcolor{blue}{\textbf{23.5}} & 20.0 & \textcolor{red}{\textbf{44.4}} & \textcolor{red}{\textbf{54.4}} & \textcolor{red}{\textbf{27.2}} & \textcolor{red}{\textbf{33.1}} & \textcolor{red}{\textbf{20.3}} & \textcolor{blue}{\textbf{17.4}} & \textcolor{red}{\textbf{33.5}} & \textcolor{blue}{\textbf{14.0}} & \textcolor{red}{\textbf{53.4}} & \textcolor{red}{\textbf{72.2}} & \textcolor{blue}{\textbf{40.9}} & {10.6} & 22.6 & \textcolor{red}{\textbf{23.1}} & \textcolor{red}{\textbf{41.5}} & \textcolor{blue}{\textbf{46.0}} & \textcolor{blue}{\textbf{44.1}} & \textcolor{blue}{\textbf{36.8}}\\

 \bottomrule
\end{tabular}

%% file: tabs/tab2.tex
\begin{tabular}{ccccccc}
\toprule
KG&Method & 2in & 3in & inp & pin & $\text{A}_{\text{N}}$ \\ \midrule
\multirow{5}{*}{FB15K-237} & ConE & 5.4 & 8.6 & 7.8 & 4.0 & 5.9 \\
&LMPNN & 8.7 & 12.9 & 7.7 & 4.6 & 8.5 \\ 
&GNN-QE & 10.0 & 16.8 & 9.3 & 7.2 & 8.5 \\
&FIT/QTO & \textcolor{red}{\textbf{14.0}} & \textcolor{blue}{\textbf{20.0}} & \textcolor{red}{\textbf{10.2}} & \textcolor{red}{\textbf{9.5}} & \textcolor{red}{\textbf{13.4}} \\
&NLISA (Local) & 12.8 & 17.3 & \textcolor{blue}{\textbf{9.5}} & 9.0 & 12.2 \\
&NLISA (Global) & \textcolor{blue}{\textbf{13.8}} & \textcolor{red}{\textbf{20.1}} & \textcolor{blue}{\textbf{9.5}} & \textcolor{blue}{\textbf{9.3}} & \textcolor{blue}{\textbf{13.2}} \\ \midrule

\multirow{5}{*}{FB15K} & ConE & 17.9 & 18.7 & 12.5 & 9.8 & 14.8 \\
&LMPNN & 29.1 & 29.4 & 14.9 & 10.2 & 20.9 \\ 
&GNN-QE & \textcolor{blue}{\textbf{44.7}} & \textcolor{blue}{\textbf{41.7}} & \textcolor{red}{\textbf{42.0}} & 30.1 & \textcolor{blue}{\textbf{39.6}} \\
&FIT/QTO & 40.2 & 38.9 & 34.8 & 28.1 & 35.5 \\
&NLISA (local) & 44.8 & 39.9 & 36.8 & \textcolor{blue}{\textbf{33.0}} & 38.6 \\
&NLISA (Global) & \textcolor{red}{\textbf{48.3}} & \textcolor{red}{\textbf{44.4}} & \textcolor{blue}{\textbf{38.0}} & \textcolor{red}{\textbf{34.5}} & \textcolor{red}{\textbf{41.3}} \\ \midrule

\multirow{5}{*}{NELL} &ConE & 5.7 & 8.1 & 10.8 & 3.5 & 6.4 \\
&LMPNN & 8.5 & 10.8 & 12.2 & 3.9 & 8.9 \\ 
&GNN-QE & 9.9 & 14.6 & 11.4 & 6.3 & 10.6 \\
&FIT/QTO & \textcolor{red}{\textbf{12.6}} & \textcolor{blue}{\textbf{16.4}} & \textcolor{red}{\textbf{15.3}} & \textcolor{red}{\textbf{8.3}} & \textcolor{red}{\textbf{13.2}} \\
&NLISA (Local) & 12.7 & \textcolor{red}{\textbf{16.5}} & \textcolor{blue}{\textbf{15.2}} & \textcolor{red}{\textbf{8.3}} & \textcolor{red}{\textbf{13.2}} \\
&NLISA (Global) & \textcolor{blue}{\textbf{12.5}} & \textcolor{red}{\textbf{16.5}} & \textcolor{red}{\textbf{15.3}} & \textcolor{red}{\textbf{8.3}} & \textcolor{red}{\textbf{13.2}} \\ \bottomrule
\end{tabular} 

%% file: cc_codex/NLISA_arxiv/NLISA_appendix.tex
\section{Experimental Details}\label{app: implementation}

\subsection{Implementation}
All experiments were conducted on a single NVIDIA V100 GPU with 32 GB of memory. Since symbolic search methods do not involve stochastic uncertainty, all results were obtained from single-run evaluations.

\paragraph{Complex query answering} For the FB15K, FB15K-237, and NELL datasets, we utilize the checkpoints provided by CQD-CO~\citep{arakelyan_complex_2020}.  Since FIT is equivalent to QTO in tree-form queries~\citep{yin_rethinking_2023}, we do not distinguish them in this case.

\paragraph{Knowledge graph completion} We reproduce the results from previous work \citep{chen2021relation} to train the embedding and hyper-network. For the FB400K dataset, we search the hyperparameters with the following settings: learning rates of $[1 \times 10^{-1}, 1 \times 10^{-2}]$, embedding dimensions of $[100, 200, 400]$, and $\lambda$ values of $[0.0005, 0.005, 0.01, 0.05, 0.1, 0.5, 1, 0]$. We utilize the ComplEx model with the N3 regularizer. The embedding initialization is set to $1 \times 10^{-3}$, and we employ the Adagrad optimizer.

\subsection{Dataset statistics}\label{app:details kg }

We follow the license of the involved KGs, BetaE dataset~\citep{ren_beta_2020}, and real EFO1 dataset~\citep{yin_rethinking_2023}, where our usage of these datasets is consistent with their intended use. The statistics of the four involved KGs are reported in Table~\ref{tab:KGs}, including entities, relations, edges, and train/validation/test splittings. 

\subsection{Query types}\label{app:details query types}

\begin{figure*}[t]
    \centering
    \includegraphics[width=0.8\linewidth]{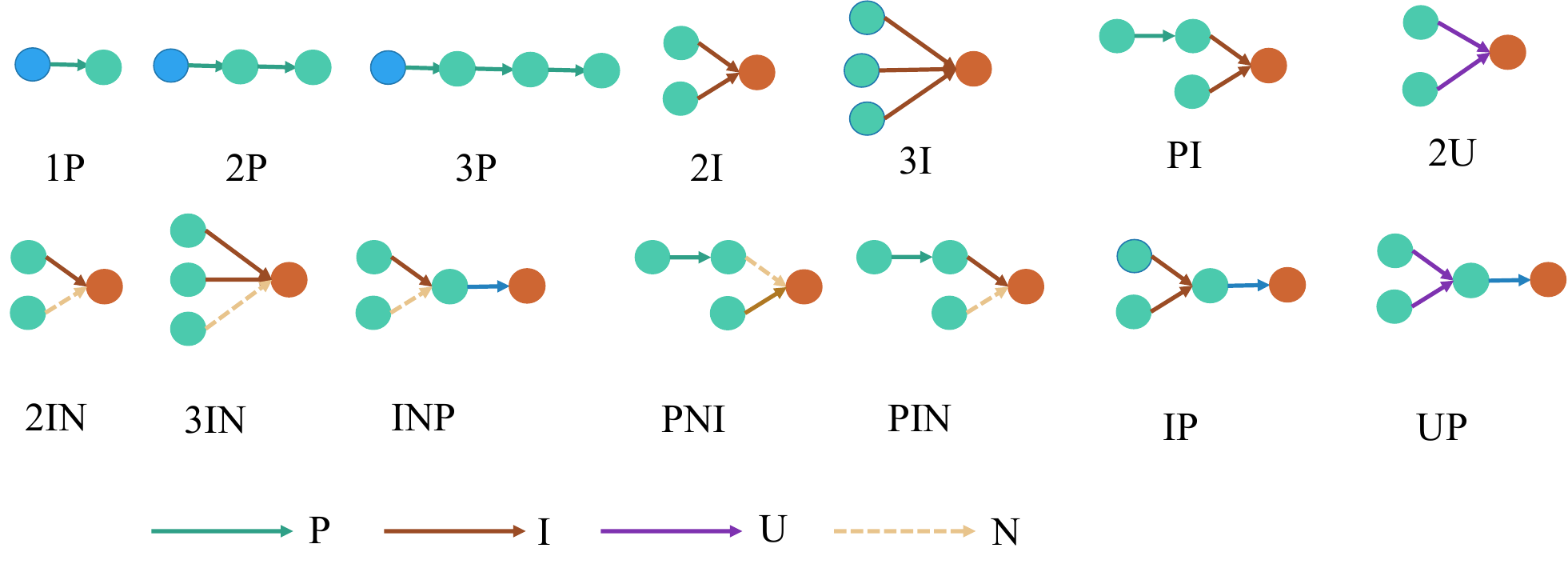}
    \caption{14 query types proposed in BetaE benchmark~\citep{ren_beta_2020}. These query types are modeled by the operator tree.}\label{fig:query types betaes}
\end{figure*}
\begin{figure*}[t]
    \centering
    \includegraphics[width=0.8\linewidth]{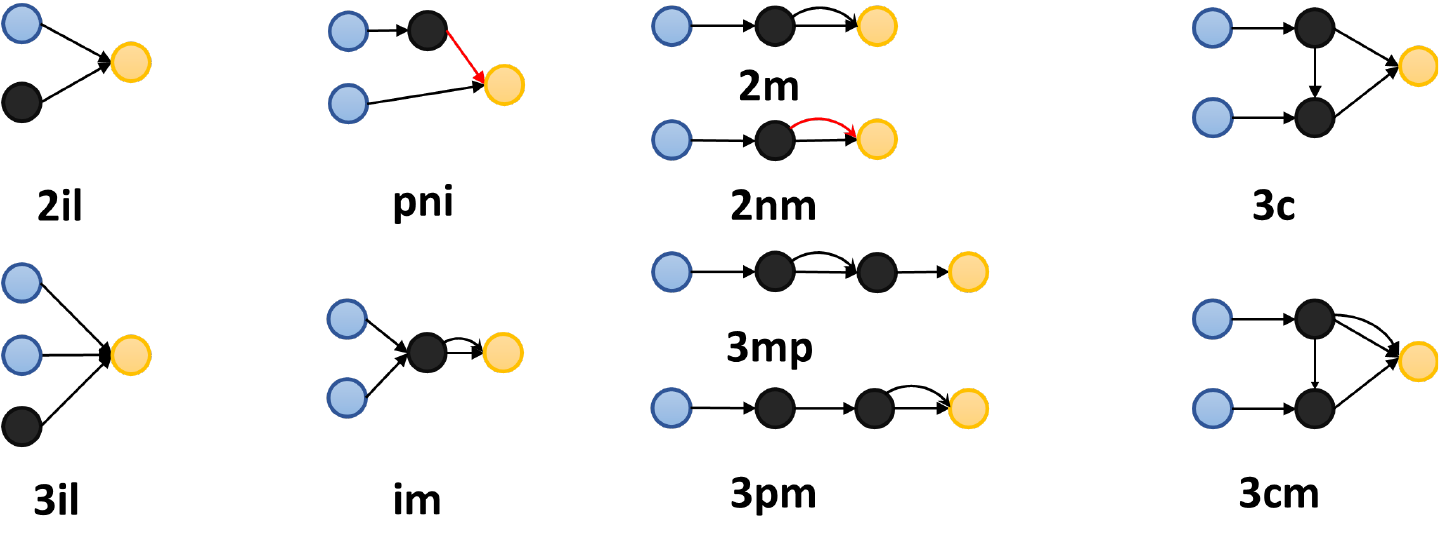}
    \caption{10 query types proposed in real EFO1 benchmark~\citep{yin_rethinking_2023}. These query types are modeled by query graph.}\label{fig:query types real efo1}
\end{figure*}

The study of complex query answering (CQA) aims to support query types with increasingly complex structures.
CQA queries are composed of nodes and edges combined with different logical operators, such as conjunction, disjunction, negation, and quantification. As a result, datasets proposed for CQA typically introduce query types with richer structural and logical properties.
The classical BetaE dataset ~\citep{ren_beta_2020} includes path queries (1p, 2p, 3p) to model relational composition, conjunctive queries (2i, 3i, ip, pi), as well as conjunctive queries with negation (2in,3in,inp,pni,pin) and disjunctive queries (2u, up). Building upon this, the real EFO1 dataset~\citep{yin_rethinking_2023}  further extends the query space by relaxing the tree-structured assumption, introducing cyclic queries (3c, 3cm), queries without constant node(2il, 3il), multi-edge queries(2m, im, 2nm, 3mp, 3pm), and refined query types (pni).
We present the visualizations of the query types for the BetaE benchmark  and the real EFO1 benchmark \citep{yin_rethinking_2023} in Figure~\ref{fig:query types betaes} and Figure~\ref{fig:query types real efo1}, respectively. The visualization for the BetaE benchmark is sourced from \citet{wang2023wasserstein}, while the visualization for the real EFO1 benchmark is derived from \citet{yin_rethinking_2023}.

\section{Why Domain Pruning Works}\label{app:theory}

In this section, we provide an explanatory analysis that helps interpret why domain reduction can work well in practice and how its behavior scales with key factors. We emphasize upfront that the following analysis relies on simplifying assumptions and should be understood as a modeling abstraction rather than a formal guarantee.

\begin{figure*}[ht]
    \centering
    \begin{minipage}{0.33\textwidth}
        \centering
        \includegraphics[width=\linewidth]{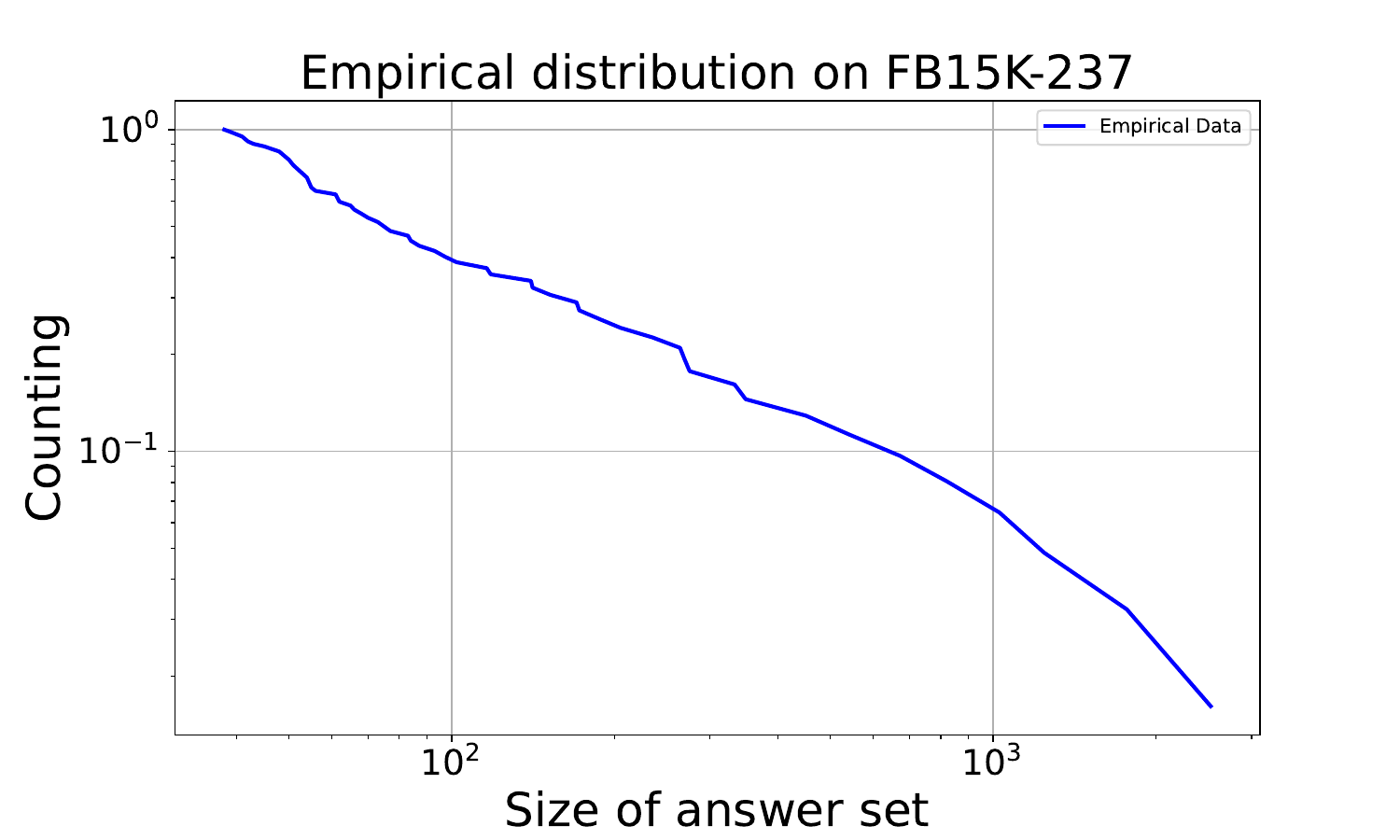}
    \end{minipage}\hfill
    \begin{minipage}{0.33\textwidth}
        \centering
        \includegraphics[width=\linewidth]{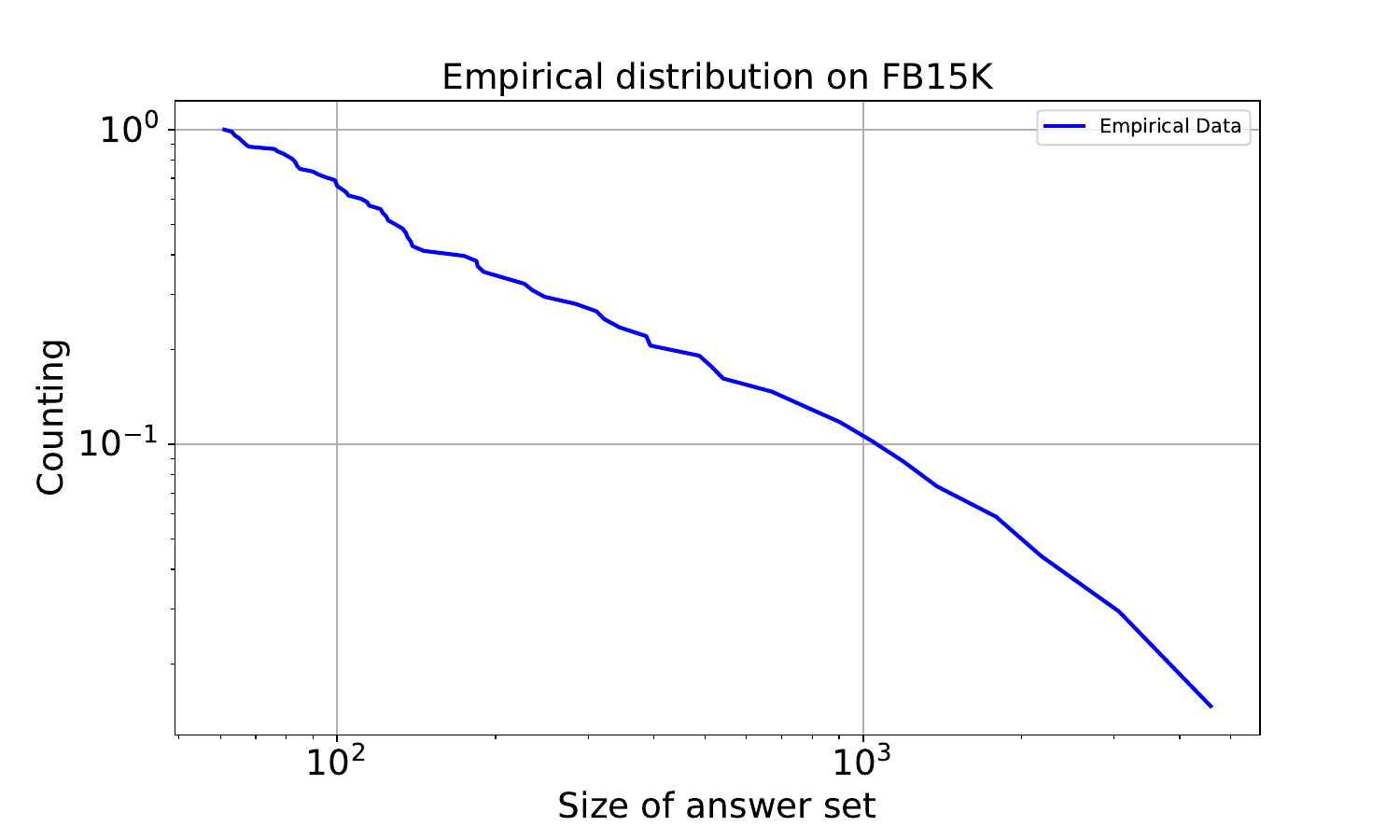}
    \end{minipage}\hfill
    \begin{minipage}{0.33\textwidth}
        \centering
        \includegraphics[width=\linewidth]{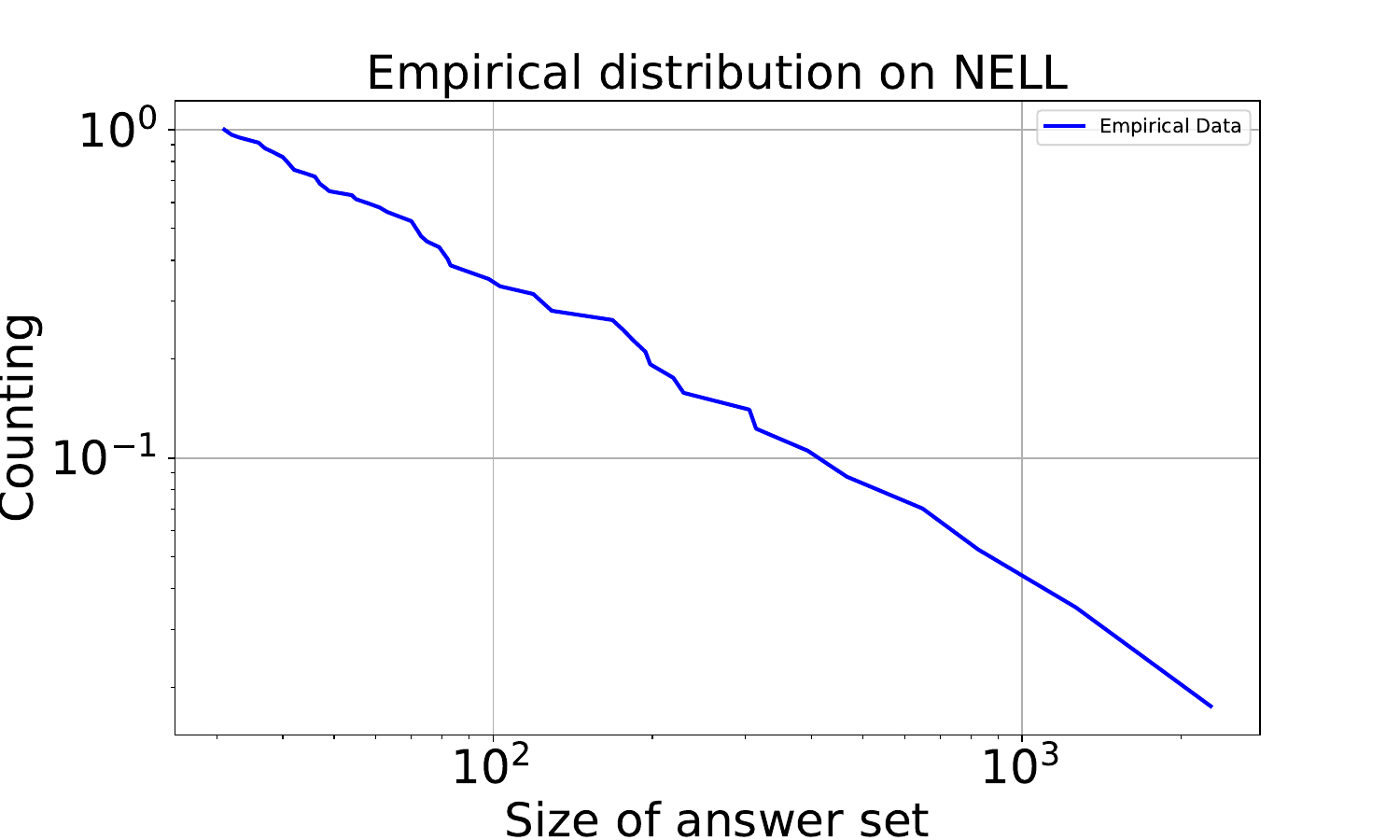}
    \end{minipage}
    \caption{Empirical distributions of answer-set/domain sizes across three knowledge graphs (KGs). Since the distributions resemble power laws, we use logarithmically spaced bins and log--log axes. The near-linearity suggests heavy-tailed behavior.}
    \label{fig:empirical-distributions}
\end{figure*}

\paragraph{Working assumption.}
Empirically, the (answer-set) domain sizes of variables exhibit heavy-tailed behavior, where we provide the results of the hypothesis testing for a power-law distribution in Figure~\ref{fig:empirical-distributions} and Table~\ref{tab:pvalues}). Motivated by this observation, we model the domain size $K$ of a variable as a random variable drawn from a discrete power-law distribution on $\{1,\dots,E\}$ with exponent $\alpha>1$. Under this model, most variables have relatively small domains, while very large domains occur infrequently.

\paragraph{Single-variable preservation (idealized setting).}
Suppose we retain the top $q$-fraction of entities (e.g., $q=0.1$ keeps 10\%). If the true domain size $K$ does not exceed $qE$ and the retained entities are ranked consistently with true membership,\footnote{This ``oracle ranking'' assumption simplifies the analysis. In practice, neural scores are imperfect but empirically correlated with membership, which partially justifies this abstraction.} then all true answers for that variable are preserved. The probability of this event is
\begin{equation}
P_{\mathrm{A}}(q;E,\alpha)
= \Pr[K \le qE]
= \frac{\sum_{k=1}^{\lfloor qE\rfloor} k^{-\alpha}}{\sum_{k=1}^{E} k^{-\alpha}}.
\label{eq:single}
\end{equation}
As $E$ grows and $\alpha>1$, most probability mass concentrates on small $k$, so $P_{\mathrm{A}}$ increases rapidly with $q$.
This explains why even modest reductions can retain a large fraction of valid answers in practice.

\paragraph{Multiple variables (independence as a proxy).}
Let $\tau$ denote the number of variables in a conjunctive query. In real queries, variable domains are often correlated through shared constraints. Analyzing such dependencies precisely is intractable in general. As a \emph{conservative and interpretable proxy}, we ignore higher-order correlations and approximate the overall preservation probability by the product of single-variable events:
\begin{equation}
P_{\mathrm{A}}^{(\tau)}(q;E,\alpha)
= \big(P_{\mathrm{A}}(q;E,\alpha)\big)^{\tau}.
\label{eq:multi}
\end{equation}
We stress that this expression should \emph{not} be interpreted as a formal guarantee. Instead, it provides a qualitative description of how query complexity ($\tau$) compounds the risk of aggressive pruning and why slightly larger $q$ may be needed for more complex queries.

\paragraph{Limitations and interpretation.}
This analysis has two clear limitations.
First, the power-law assumption is empirically motivated rather than theoretically guaranteed. Second, the independence approximation ignores correlations induced by query structure. Consequently, the results should be viewed as an \emph{intuition-building model} rather than a correctness proof.

Nevertheless, the analysis offers useful explanatory insights:
(i) heavy-tailed domain sizes make moderate pruning effective,
(ii) preservation degrades smoothly as pruning becomes more aggressive, and
(iii) query complexity primarily affects preservation through the number of variables.
These trends are consistent with our empirical results, where retaining about $10\%$ of entities typically preserves over $95\%$ of answers across datasets.

\begin{table}[t]
    \centering
    \caption{Power-law fits for domain-size distributions on three KGs. We report the fitted exponent $\alpha$ and the Kolmogorov--Smirnov (KS) $p$-value (higher is better).}
    \label{tab:pvalues}
    \begin{tabular}{lccccc}
    \toprule
    & \multicolumn{1}{c}{FB15K-237} & \multicolumn{1}{c}{FB15K} & \multicolumn{1}{c}{NELL} \\
    \midrule
    Fitted $\alpha$      & 1.93 & 1.89 & 1.92 \\
    KS $p$-value         & 0.84 & 0.63 & 0.65 \\
    \bottomrule
    \end{tabular}
\end{table}

\section{Fuzzy Logical Inference}\label{app: fuzzy logic}
\subsection{$t$-norms}\label{app:t-norm introduction}
\begin{definition}[$t$-norm]
    A $t$-norm $\top$ is a function: [0,1] x [0,1] $\rightarrow$ [0,1] that satisfies the following properties:
    \begin{itemize}
        \item[(i)] Commutativity: $a\top b = b\top a$
        \item[(ii)] Monotonicity: $(a\top b) \leq  (c\top b)$ if $a\leq c$
        \item[(iii)] Associativity: $(a\top b)\top c= a\top (b\top c)$
        \item[(iv)] Neutrality: $a\top 1 =a$ 
    \end{itemize}
\end{definition}

Then the $t$-conorm $\bot$ is directly defined by $a\bot b = 1 - (1-a)\top(1-b)$, which follows the De Morgan's law.

Finally, we introduce some common $t$-norms which are of interest:

\begin{itemize}
    \item[(i)]  Godel: $a\top_{G} b = \text{min}(a,b)$
    \item[(ii)] Product: $a\top_{P} b = a \times b$
    \item[(iii)] Łukasiewicz: $a\top_{LK} b = \text{max}(a+b-1,0)$
\end{itemize}

In the main paper, we mainly focus on Product $t$-norm.

\subsection{Truth-value construction}
The symbolic representation in the previous symbolic search method QTO and FIT is constructing the set of truth values matrix for the whole knowledge graph. To convert real number scores computed by knowledge graph embedding modes to truth value that falls into $[0, 1]$, QTO/FIT use the softmax function:
\begin{equation}
    P_{r,a}^{\star}(b) = \frac{exp(s(a,r,b))}{\Sigma_{c\in \mathcal{E}} exp(s(a,r,c))}.
\end{equation}
Next, QTO and FIT scale the results of the softmax function using a factor based on the observed edges in the training graph since softmax outputs a vector that sums to 1:
$\mathcal{G}_o$. 
\begin{equation}
    Q_{a,b}= 
\begin{cases}
    \frac{|\{d|(a,r,d)\in\mathcal{G}_o\}|}{\Sigma_{c\in \{d|(a,r,d)\in\mathcal{G}_o\}} P_{r,a}^{\star}(c)},& \text{if } C_a^r> 0\\
    1,              & \text{if } C_a^r= 0
\end{cases}
\end{equation}
where $C_a^r = |\{d|(a,r,d)\in\mathcal{G}_o\}|$ and $E_{a,r} = \{b\mid (a,r,b)\in \mathcal{G}_o\}$ represents the set of observed edges. Then the $a$-th row of $r$-th matrix is got by clamping the value for each element:
\begin{equation}
    P_{r}(a,b)= min(1,P_{r,a}^{\star}(b) \times Q_{a,b})
\end{equation}  
They then mark the observed edges and set the truth value for these edges to 1. The scaling and marking operations are performed on a case-by-case basis for each fact, which cannot be parallelized.

We demonstrate that these scaling operations can be parallelized through caching. For cases where $C_a^r> 0$: the truth value is computed by the following normalization:
\begin{equation}
    Q_{a,b}= \frac{exp(f_r(a,b))}{\sum_{\{d|(a,r,d)\in\mathcal{G}_o\}} exp(f_r(a,d))}.
\end{equation}
To simplify the calculations, we use log-scale operations to compute $Q_{a,b}$. If we cache $S^r_{a} = log(\sum_{\{d|(a,r,d)\in\mathcal{G}_o\}} exp(f_r(a,d)))$ with $|\mathcal{E}| \times |\mathcal{R}|$ size, we can parallel index the cached values to optimize the computation of $Q_{a,b}$. For cases where $C_a^r> 0$, we have: 
\begin{equation}
    S^{r}_{a} = log(\sum_{\{d \in \mathcal{E} \}} exp(f_r(a,d))),
\end{equation}
which allows the scaling operation to yield the softmax result. 

This caching mechanism grows linearly with the size of the knowledge graph. By utilizing this caching strategy, we can perform parallel computations for the scaling operations, facilitating efficient dynamic symbolic representation construction. However, when it comes to listing the marked training facts, our symbolic representation may exhibit slightly lower performance compared to the original construction.

\subsection{Node-removal operations}\label{app:details functions}

$\textsc{RemovEConstNode}( G_{\phi},|\mathcal{D}_x|, |\mathcal{D}_y|)$ removes the constant nodes based on the given  domain of variables. We consider the simplest case where   $(s,r,e,\textsc{Neg}=\textsc{False})$, with $s$ as the constant node, $r$ as the relation, $e$ as the existential variable, and $\textsc{Neg}=\textsc{False}$ indicating that this edge is positive. We first construct the symbolic representation of this edge: $\vs = [P_r(s,e^i)] \in \R^{|\mathcal{D}_e|}, e^i \in \mathcal{D}_e$. We then update this representation into the fuzzy vector of the existential variable  $e$ by t-norm:
$\mu(e,\mathcal{D}) = \mu(e,\mathcal{D}) \top \vs $. Consequently, we can remove this edge while preserving its constraints in the fuzzy vector of $e$.  The above method can be generalized to other cases, such as when the variable is a free variable $y$ and the edge is negative. A similar update approach can be used, and specific details can be found in \citet{yin_rethinking_2023}.

$\textsc{RemoveConstNode}( G_{\phi},|\mathcal{D}_x|, |\mathcal{D}_y|)$ removes one leaf node based on the given domain of variables. We consider the most difficult case where   $(e_1,r,e_2,\textsc{Neg}=\textsc{False})$, with $s$ as the constant node, $r$ as the relation, $e$ as the existential variable, and $\textsc{Neg}=\textsc{False}$ indicating that this edge is positive. The symbolic representation of this edge results in a matrix: $\mS = [P_r(e^i_1,e^j_2)] \in \R^{|\mathcal{D}_{e_1}|\times |\mathcal{D}_{e_2}|}, e_1^i \in \mathcal{D}_{e_1}, e_2^i \in \mathcal{D}_{e_2}$.  We then update this representation into the fuzzy vector of the existential variable  $e$ by t-norm and max operation: $\mu(e,\mathcal{D}) = \mu(e,\mathcal{D}) \top \max ( \mS, axis=0) $. The proof of the effectiveness of the above update can be found in \citet{yin_rethinking_2023}, and this update can be easily extended to other cases, such as when the variable is a free variable $y$ and the edge is negative.

\section{Neural Logical Index Construction}\label{app:details strategy}
For local constraints, when the variable involves logical disjunction, we use the T-conorm to handle it. For global constraints, the computation of existential variables should treat them as free variables, while the original free variables are converted into existential variables. This allows us to use query embedding to compute the global constraints for the existential variables.

\subsection{Knowledge graph embedding and HyperNet}
Knowledge graph embedding first embeds the entities and relations into a continuous space. Given the query $(s,r,?)$, we first compute its embedding and compare it with  the entities within the embedding space. The estimated embedding of $(s,r,?)$ is first computed by $ f_t(e_s, e_r)$, where $f(\cdot,\cdot): \mathbb{R}^d \times \mathbb{R}^d \to \mathbb{R}^d$ is the transformation function.  Then, the likelihood of $r(s,o)$ is computed by the scoring function $f_s(f_t(e_r, e_{s}),e_{o})$, where $f_s(\cdot,\cdot): \mathbb{R}^d \times \mathbb{R}^d \to [-\infty, +\infty]$ is the scoring function related to the embedding space. The scoring function is usually a distance function or a similar metric in the embedding space.

For relation tail prediction task, we adapt existing KG embedding models using a Hypernet. The Hypernet modifies the embeddings of entities and relations, while we employ the same scoring function as the original embedding models. Given the query $(r,?)$ and candidate entity $t$, the score $g_s(\texttt{HN}(r),\texttt{HN}(t))$, where $\texttt{HN}$ represents the Hypernet  and $g_s = \sigma(f_s)$ with $\sigma(\dot)$ being the sigmoid activation function.

\subsection{HyperNet motivation and training}
\paragraph{Motivation}
Relation tail prediction is fundamentally similar to knowledge graph completion (KGC) tasks, as both involve inferring missing tails from a knowledge graph. In KGC, the query is presented as $(h, r, ?)$, to predict the tail $t$. In contrast, relation tail prediction aims to predict the tail $t$ given the relation $r$. This similarity makes the embedding methods used in KGC equally applicable to relation tail prediction. Hypernet is employed to generate weights dynamically, allowing the model to automatically adjust representations for new tasks. This facilitates the sharing and adaptation of weights across different tasks. By adopting the embeddings from KGC, Hypernet enhances the utilization of existing knowledge and enables quicker adaptation to relation tail prediction. Additionally, Hypernet requires only a few parameters, improving performance while reducing training costs and storage needs.

\paragraph{Training Hypernet} The training process is similar to that of KGC, where each triple $(h,r,t)$ is replaced with $(r,t)$. This requires only minor modifications compared to the existing KGC framework.

\paragraph{Performance of Hypernet} For the Hypernet used for relation tail prediction, we set the learning rate to $1 \times 10^{-2}$ and search the hidden dimensions $[100, 200, 400]$ also using the ComplEx model with the N3 regularizer. The embedding initialization remains at $1 \times 10^{-3}$, and the optimizer is Adagrad. The performance of our method is quite impressive, as demonstrated in Table~\ref{tab: hyper}. The results indicate a strong capability of the Hypernet in addressing relation tail prediction tasks.
\begin{table}[]
\centering
\caption{MRR results for the trained Hypernet.}
\label{tab: hyper}
\resizebox{0.48\textwidth}{!}{
\begin{tabular}{ccccc}
\toprule
    & FB15K-237 & FB15K & NELL & FB400K \\ \midrule
MRR & 0.95       & 0.98 & 0.95 & 0.96  \\ \bottomrule
\end{tabular}
}
\end{table}

\section{Extended Experiments}\label{app:additional result}

\subsection{Efficiency and memory usage}

\begin{table*}
\centering
\scriptsize
\caption{Efficiency results of the tree-form queries on BetaE benchmark~\citep{ren_beta_2020}.}
\label{tab: efficiency kg}
\begin{tabular}{cccccccc}
\toprule
KG                     &       Metric            & ConE & LMPNN & CQD-CO  & FIT   & NLISA (local) & NLISA (global) \\ \midrule
\multirow{3}{*}{FB15K-237} & Queries per Second~$\uparrow$   & 100  & 105   & 14   & 31  & \textbf{115}   & 86   \\
    & CUDA Memory of Running (M)~$\downarrow$ & 458  & 355   & \textbf{176}  & 1208  & 232  & 337  \\ \midrule
\multirow{3}{*}{FB15K}     & Queries per Second~$\uparrow$  &    97  & 101       &  10.1    &  20     &          \textbf{109}  &     79        \\
                           & CUDA Memory of Running (M)~$\downarrow$ &  338    &   365    &  \textbf{189}    &  2068     &   255         &   381  \\ \midrule
\multirow{3}{*}{NELL}      & Queries per Second~$\uparrow$  & 23   & 24    & 7    & 3  & \textbf{35}  & 20\\

                           & CUDA Memory of Running (M)~$\downarrow$ & 1635 & 1955  & 760 & 15324 & \textbf{774} &953       \\ \bottomrule        
\end{tabular}
 \vspace{-1em}

\end{table*}

Table~\ref{tab: efficiency kg} further reports the runtime memory usage. Compared with FIT, NLISA (L) uses 5--20$\times$ less CUDA memory across FB15K-237, FB15K, and NELL. The saving mainly comes from activation memory during search: FIT operates on full entity-level intermediate states, whereas NLISA performs node removal and local optimization over pruned domains of size $|\mathcal{E}|/\kappa$ per variable. The parameter memory of the KGE embeddings and lightweight neural components is shared with the backbone and is not the main source of the gap.

\subsection{Recent query embedding baselines}

We further discuss two recent query embedding baselines, Query2Triple and context-aware query representation learning. These methods focus on tree-structured queries and are not designed for cyclic EFO1 queries, so we compare them on the positive and negative query groups of the BetaE benchmark. The results are summarized in Table~\ref{tab:recent_qe}. NLISA (G) is competitive on positive queries and shows clear advantages on negative queries, where pruning removes many low-relevance candidates before symbolic ranking.

\begin{table}[t]
\centering
\caption{Comparison with recent query embedding methods on positive (AP) and negative (AN) tree-form query groups.}
\label{tab:recent_qe}
\resizebox{0.35\textwidth}{!}{
\begin{tabular}{llcc}
\toprule
Dataset & Method & AP & AN \\
\midrule
\multirow{4}{*}{FB15K-237}
& Query2Triple & 27.6 & 7.85 \\
& ConE+CaQR & 23.67 & 6.43 \\
& NLISA (L) & 25.2 & 12.2 \\
& NLISA (G) & 27.0 & 13.2 \\
\midrule
\multirow{4}{*}{NELL}
& Query2Triple & 31.5 & 7.08 \\
& ConE+CaQR & 27.35 & 7.19 \\
& NLISA (L) & 33.2 & 13.2 \\
& NLISA (G) & 33.5 & 13.2 \\
\bottomrule
\end{tabular}
}
\end{table}

\subsection{Domain-size sensitivity}
\begin{table}[t]
\centering
\caption{Ablation study under different domain sizes.}
\label{tab:aba_domain_size_results}
\small
\begin{tabular}{l l l c c c c c}
\toprule
\textbf{Dataset} & \textbf{Method} & \textbf{Metric}
& \textbf{1000} & \textbf{2000} & \textbf{4000} & \textbf{6000} & \textbf{8000} \\
\midrule
\multirow{5}{*}{FB15K-237}
& \multirow{3}{*}{BetaE}
& $\kappa$ & 14.505 & 7.2525 & 3.62625 & 2.4175 & 1.813125 \\
&  & Times  & 382.5  & 454.2  & 1390.9  & 2024.8 & 3305.4  \\
&  & MRR    & 23.8   & 25.4   & 26.4    & 26.6   & 26.6   \\
& \multirow{2}{*}{real EFO1}
& Times  & 389.3  & 464.2  & 1625.6  & 2688.5 & 5235.2 \\
&  & MRR    & 20.7   & 22.2   & 22.9    & 23.0   & 23.6   \\
\midrule
\multirow{3}{*}{FB15K}
& \multirow{3}{*}{BetaE}
& $\kappa$ & 14.951 & 7.4755 & 3.73775 & 2.491833333 & 1.868875 \\
&  & Times  & 2232.4 & 3197.2 & 4177.5 & 4188.2 & 6720.2 \\
&  & MRR    & 59.2   & 71.1   & 71.3   & 71.9   & 72.1   \\
\midrule
\multirow{7}{*}{NELL}
& \multirow{2}{*}{real EFO1}
& Times  & 433.0  & 790.5  & 1736.8 & 2052.3 & 3464.8 \\
&  & MRR    & 50.8   & 62.0   & 63.6   & 64.0   & 64.1   \\
& \multirow{3}{*}{BetaE}
& $\kappa$ & 63.361 & 31.6805 & 15.84025 & 10.56016667 & 7.920125 \\
&  & Times  & 2820.8 & 2822.9 & 2856.6 & 3263.2 & 3944.2 \\
&  & MRR    & 33.6   & 33.8   & 33.9   & 33.9   & 33.9   \\
& \multirow{2}{*}{real EFO1}
& Times  & 627.1  & 1154.3 & 1823.6 & 2874.1 & 5376.1 \\
&  & MRR    & 32.9   & 36.8   & 37.1   & 37.3   & 37.4   \\
\bottomrule
\end{tabular}
\end{table}

We provide the detailed ablation study in Table~\ref{tab:aba_domain_size_results}, which includes the results on different KGs and different benchmarks.

\subsection{KGE backbone sensitivity}

\begin{table*}[t]
\centering
\caption{Performance comparison on FB15K-237 and FB15K under different KGE backbone.}\label{tab:ab_kge_backbone}

\small
\setlength{\tabcolsep}{4pt}
\begin{tabular}{l l c c c c c c c c c c c c c c c}
\toprule
\textbf{Dataset} & \textbf{Method} 
& 1p & 2p & 3p & 2i & 3i & ip & pi & 2u & up & 2in & 3in & inp & pin & $A_p$ & $A_N$ \\
\midrule
\multirow{2}{*}{FB15K-237}
& Base 
& 46.6 & 14.3 & 12.3 & 36.2 & 50.0 & 21.2 & 29.3 & 17.6 & 12.5 & 12.4 & 17.1 & 9.3 & 8.3 & 26.67 & 11.78 \\
& QTO 
& 48.6 & 19.6 & 19.8 & 42.2 & 56.3 & 26.5 & 34.2 & 19.7 & 18.0 & 14.0 & 21.7 & 12.5 & 12.8 & 31.66 & 15.25 \\
\midrule
\multirow{2}{*}{FB15K}
& Base 
& 89.4 & 64.7 & 55.4 & 76.4 & 82.3 & 71.1 & 71.1 & 71.9 & 57.5 & 48.3 & 44.4 & 37.9 & 34.5 & 71.09 & 41.28 \\
& QTO 
& 89.3 & 64.2 & 53.9 & 75.7 & 81.9 & 71.4 & 69.8 & 71.1 & 57.3 & 55.0 & 48.6 & 40.8 & 36.4 & 70.51 & 45.20 \\
\bottomrule
\end{tabular}
\end{table*}

For a fair comparison, we adopt the same KGE backbone for all methods. In addition, we report the results of NLISA (G) using another widely adopted backbone~\citep{bai_answering_2023}, which incorporates relation prediction as an auxiliary training objective~\citep{chen_learning_2023}. We denote this alternative backbone as ``QTO'', following its first introduction, and refer to the backbone used in the main paper as ``Base''. The results on FB15K-237 and FB15K are presented in Table~\ref{tab:ab_kge_backbone}.  As shown in the table, the new backbone further improves overall performance. The gains are particularly pronounced on FB15K-237.

\subsection{Hard-query benchmark}

\begin{table*}[!t]
\center
\setlength{\tabcolsep}{5.5pt}
\caption{Results on new benchmark~\citep{gregucci2024complex}. The results of all the baselines are sourced from \citet{gregucci2024complex}. The best results over all the models are highlighted in black.}\label{tab:new-bench+h-MRR}
\scalebox{.88}{
\vspace{3pt}
\begin{tabular}{crcccccccccccccc}
\toprule
& Model & 1p & 2p & 3p & 2i & 3i & 1p2i & 2i1p &2u& 2u1p& 2in& 3in& 2pi1pn& 2nu1p& 2in1p\\
\midrule
\multirow{6}{*} &
  GNN-QE & 42.8 & 5.2 & 4.0 & 6.0 & 8.8 & 5.6 & 9.9 & 32.5 & 10.0 & 6.8&\textbf{6.5}&\textbf{3.7}&5.0&3.3\\ 
  & ULTRAQ& 40.6 & 4.5 & 3.5 & 5.2 & 7.2 & 5.3 & 10.1 & 29.4 & 8.3 &5.3&5.5&2.6&3.7&2.2\\ 
  &CQD & \textbf{46.7} & 4.4 & 2.4 & 11.3 & \textbf{12.8} & 6.0 & 11.5 & 40.1 & 10.6 &3.3&2.6&0.6&4.9&1.2\\ 
  &CQD-Hybrid & \textbf{46.7} & 4.8 & 3.1 & 6.0 & 8.6 & 5.5 & 12.9 & \textbf{42.2} & 12.0 &4.7&1.6&1.0&3.2&1.3\\ 
  &ConE & 41.8 & 4.6 & 3.9 & 9.1 & 10.3 & 3.8 & 7.9 & 22.8 & 6.0 &5.1&4.9&2.9&3.3&\textbf{3.6}\\ 
  &QTO & \textbf{46.7} & 4.9 & 3.7 & 8.7 & 10.1 & 6.1 & 13.5 & 30.6 & 11.2 &10.6&3.1&2.0&\textbf{5.3}&1.5\\ 
  &CLMPT & 45.3 & \textbf{5.3} & \textbf{4.7} & 10.2 & 12.2 & 5.6 & \textbf{14.9} & 33.6 & \textbf{14.2}  &6.8&2.3&1.6&4.8&2.5\\ 
  &NLISA (G) & 46.6 & \textbf{5.3} & 3.6 & \textbf{11.9} & \textbf{13.2} & \textbf{6.6} & 12.6 & 34.7 & 10.1 &\textbf{13.8}&3.9&1.6&{4.4}&2.2\\ 
\midrule
\multirow{6}{*}
&GNN-QE& 53.6 & 8.0 & 6.0 & 10.7 & 13.3 & 16.0 & 13.5 & 47.5 & 9.8 
& 5.5 & \textbf{6.4} & \textbf{5.8} & 3.3 & 4.4\\ 
&ULTRAQ& 38.9 & 6.1 & 4.1 & 7.9 & 10.2 & 15.8 & 9.3 & 28.1 & 9.5 
& 4.5 & 5.9 & 4.3 & 2.7 & 3.6\\ 
&CQD& {60.4} & 9.6 & 4.2 & 18.5 & 19.6 & 18.9 & 22.6 & 46.3 & 18.5 
& 4.2 & 1.5 & 1.5 & 4.9 & 2.6\\ 
&CQD-Hybrid& {60.4} & 9.0 & 6.1 & 12.1 & 14.4 & 17.4 & 21.2 & 46.4 & 19.3 
& 5.1 & 1.2 & 1.4 & 4.3 & 2.4\\ 
&ConE& 53.1 & 7.9 & 6.7 & 21.8 & 23.6 & 14.9 & 11.8 & 39.9 & 8.8& 4.6 & 6.0 & 3.7 & 2.7 & \textbf{6.4}\\ 
&QTO & 60.3 & 9.8 & \textbf{8.0} & 14.6 & 15.8 & {17.6} & 21.1 & 49.1 & 18.9 
& {10.2} & 2.3 & 3.1 & \textbf{8.4} & 2.4\\ 
&CLMPT & 58.1 & {10.1} & 7.8 & \textbf{22.7} & \textbf{25.0} & 17.2 & \textbf{24.4} & {50.0} & \textbf{22.0} & 6.5 & 2.4 & 4.1 & 2.3 & 4.5\\ 
&NLISA (G) & \textbf{60.8} & \textbf{10.3} & 7.6 & {19.1} & {20.1} & \textbf{18.7} & {23.1} & \textbf{50.1} & {18.2} & \textbf{12.5} & 2.6 & 2.5 & 5.0 & 2.5\\ 
  \bottomrule
\end{tabular}
}
\end{table*}
We further conduct experiments with NLISA on FB15K-237+H and NELL+H, which were proposed to evaluate hard queries where all reasoning paths to the correct answers contain at least one incomplete edge.
The results in Table~\ref{tab:new-bench+h-MRR} show that NLISA achieves competitive performance: NLISA (G) attains the best MRR on 5 out of 14 query types among all baselines.

\section{NLISA Algorithms}\label{sec: algorithm of nlisa}
\begin{algorithm}[t]
\caption{NLISA: Neural Logical Indices for Search Approximately}
\label{alg:nlisa_appendix}
\begin{algorithmic}[1]
\REQUIRE Conjunctive query $\phi$ with free variable $y$; pretrained KG embeddings; reduction factor $\kappa$.
\ENSURE Answer scores $\{\phi\mid_{y=s}\}_{s \in \mathcal{E}}$.

\STATE \textbf{Step 1: Build Neural Logical Indices for prune domains}
\FOR{each variable $x$ in $\phi$}
    \STATE Compute fuzzy membership scores $C_x[e]$ for all $e \in \mathcal{E}$ using the embedding model based on subgraph $G^S_x$.
    \STATE Sort entities $e$ in descending order of $C_x[e]$.
    \STATE Let $\mathcal{D}_x$ be the top-$\lfloor |\mathcal{E}|/\kappa \rfloor$ entities (reduced domain for $x$).
\ENDFOR

\STATE \textbf{Step 2: Constant node elimination}
\STATE Let $\psi \leftarrow \phi$.
\WHILE{$\psi$ contains a constant node }
    \STATE Remove this node and its incident edges from $\psi$.
    \STATE Propagate the removed constraints into the remaining fuzzy vectors $\{C_x\}$.
\ENDWHILE

\STATE \textbf{Step 3: Leaf-node elimination}
\STATE Let $\psi \leftarrow \phi$.
\WHILE{$\psi$ contains a leaf variable node}
    \STATE Remove this node and its incident edges from $\psi$.
    \STATE Propagate the removed constraints into the remaining fuzzy vectors $\{C_x\}$.
\ENDWHILE

\STATE \textbf{Step 4: Solve the cyclic query}
\IF{$\psi$ is acyclic}
    \STATE Evaluate $\psi$ over reduced domains $\{\mathcal{D}_y\}$.
\ELSE
    \STATE Call \textsc{LOCALOPTIMIZE}$(\psi, \{\mathcal{D}_x\}, \{C_x\}, y)$ to obtain scores $\{\phi\mid_{y=s}\}$.
\ENDIF

\RETURN $\{\phi\mid_{y=s}\}_{s \in \mathcal{E}}$, where $\phi\mid_{y=s} = 0, {s \notin \mathcal{D}_y}$.
\end{algorithmic}
\end{algorithm}

\begin{algorithm}[t]
\caption{\textsc{LOCALOPTIMIZE} for cyclic queries}
\label{alg:localoptimize}
\begin{algorithmic}[1]
\REQUIRE Reduced cyclic query $\psi$; reduced domains $\{\mathcal{D}_x\}$; fuzzy vectors $\{C_x\}$; free variable $y$.
\ENSURE Scores $\{\psi\mid_{y=s}\}_{s \in \mathcal{D}_y}$.

\STATE Let remaining variables be $\mathcal{X}_r = \{x_1, \dots, x_m\}$, ordered by shortest-path distance to $y$ in $\psi$.

\FOR{each $s \in \mathcal{D}_y$}
    \STATE Initialize assignment $A \leftarrow \{y \mapsto s\}$.
    \FOR{$i = 1$ to $m$}
        \STATE \textbf{ greedy selection of $x_i$ from its reduced domain}
        \FOR{each $a \in \mathcal{D}_{x_i}$}
            \STATE Construct the neighborhood subgraph $G_n(x_i, \psi)$.
            \STATE Compute local constraint score $u(a)$ by aggregating $T(e)$ over all $e \in G_n(x_i,\psi)$.
            \STATE Compute membership score $v(a) \leftarrow C_{x_i}[a]$.
            \STATE Let $q(a)$ be the $t$-norm aggregation of $u(a)$ and $v(a)$.
        \ENDFOR
        \STATE Choose $x_i^s \in \arg\max_{a \in \mathcal{D}_{x_i}} q(a)$.
        \STATE Update $A \leftarrow A \cup \{x_i \mapsto x_i^s\}$.
    \ENDFOR
    \STATE Evaluate $\psi\mid_{y=s}$ by applying the truth function $T(\cdot)$ to $\psi$ instantiated with assignment $A$.
\ENDFOR

\RETURN $\{\psi\mid_{y=s}\}_{s \in \mathcal{D}_y}$.
\end{algorithmic}
\end{algorithm}

We provide the pseudo code of NLISA in Algorithm~\ref{alg:nlisa_appendix}. Additionally, we provide the detailed pseudo code for \textsc{LOCALOPTIMIZE} in Algorithm~\ref{alg:localoptimize}.

\section{Propagation Rules}
\label{app:propagation details}

In this section, we provide the detailed derivations for the propagation rules used in
Section~\ref{sec: effcient search} (including existing methods QTO~\citep{bai_answering_2023} and FIT~\citep{yin_rethinking_2023}).  We elaborate on different structural cases of logical queries following prior
neural-symbolic frameworks~\citep{yin_rethinking_2023,bai_answering_2023}.
Our goal is to derive the answer vector $\mathcal{A}(\phi)$ for conjunctive queries
$\phi(y)$. The answer vector of a general EFO1 query can then be computed via its disjunctive
normal form (DNF)~\citep{ren_query2box_2020}.

The core idea is to iteratively remove nodes and edges from the query graph while
propagating their constraints into fuzzy membership vectors, such that inference can
be carried out on progressively smaller graphs without loss of correctness.

\subsection{Step 1: Initialization}

We initialize fuzzy membership vectors for all non-entity nodes in the query graph
$G_{\phi}$. 
For each variable node $u$ (either an existential variable or the free variable), we
initialize $C_u = \mathbf{1}$ as an all-one vector.
The original query formula is equivalently rewritten as
\[
\phi(y) \land \mu(u, C_u), \quad \forall u,
\]
which explicitly encodes the initial unconstrained domains of variables.

\subsection{Step 2: Removing Self-Loops}

For any node $u$ with self-loop edges, we distinguish two cases:
(1) $u$ is the free variable $y$, and
(2) $u$ is an existential variable.
Self-loops on constant entities are ignored since they do not impose meaningful
constraints.

\noindent\textbf{Case 1: $u = y$ is the free variable.}

Assume that $y$ has a positive self-loop $r(y,y)$.
By rearranging conjuncts, the query can be written as
\begin{align*}
    \phi(y) = \mu(y,C_y) \land r(y,y) \land \psi(y),
\end{align*}
where $\psi(y)$ denotes the remaining subformula.

For any entity $a \in \mathcal{E}$, the answer score is
\begin{align*}
    A[\phi(y)](a)
    &= T(\mu(a,C_y) \land r(a,a) \land \psi(a)) \\
    &= C_y(a) \top P_r(a,a) \top T(\psi(a)).
\end{align*}

Introducing $\odot^{\top}$ as element-wise application of the $t$-norm, we obtain
\begin{align*}
    A[\phi(y)]
    &= (C_y \odot^{\top} \diag(P_r)) \odot^{\top} A[\psi(y)] \\
    &= \mu(y, C_y \odot^{\top} \diag(P_r)) \odot^{\top} A[\psi(y)].
\end{align*}

Thus, the self-loop can be removed by updating the fuzzy vector $C_y$ accordingly.

If the self-loop is negative, i.e., $\lnot r(y,y)$, the derivation is analogous:
\begin{align*}
    A[\phi(y)](a)
    &= C_y(a) \top (1 - P_r(a,a)) \top T(\psi(a)), \\
    A[\phi(y)]
    &= \mu(y, C_y \odot^{\top} (1 - \diag(P_r))) \odot^{\top} A[\psi(y)].
\end{align*}

\noindent\textbf{Case 2: $u$ is an existential variable $x$.}

Assume that $x$ has $n$ positive self-loops
$r_1(x,x), \ldots, r_n(x,x)$.
The query can be written as
\[
\phi(y) = \exists x.\,
\mu(x,C_x) \land r_1(x,x) \land \cdots \land r_n(x,x) \land \psi(y;x),
\]
where $\psi(y;x)$ contains no further self-loops on $x$.

For any $a \in \mathcal{E}$,
\begin{align*}
    T(\phi(a))
    &= \bot^{\star}_{b \in \mathcal{E}}
    [C_x(b) \top P_{r_1}(b,b) \top \cdots \top P_{r_n}(b,b) \top T(\psi(a;b))] \\
    &= \bot^{\star}_{b \in \mathcal{E}}
    [C_x'(b) \top T(\psi(a;b))],
\end{align*}
where
\[
C_x' = C_x \odot^{\top} \diag(P_{r_1}) \odot^{\top} \cdots \odot^{\top} \diag(P_{r_n}).
\]

Thus, all self-loops on $x$ can be removed by updating $C_x$ to $C_x'$.
After this step, the query graph contains no self-loops.

\subsection{Step 3: Removing Constant Entities}

If a node represents a constant entity $a$, all edges incident to $a$ can be removed.
We distinguish two cases depending on whether $a$ connects to the free variable or to
an existential variable.

\noindent\textbf{Case 1: $a$ connects to the free variable $y$.}

Assume two positive edges $r_1(a,y)$ and $r_2(y,a)$.
The query is
\[
\phi(y) = \mu(y,C_y) \land r_1(a,y) \land r_2(y,a) \land \psi(y).
\]

For any $b \in \mathcal{E}$,
\begin{align*}
    A[\phi(y)](b)
    &= C_y(b) \top P_{r_1}(a,b) \top P_{r_2}^{\intercal}(a,b) \top T(\psi(b)) \\
    &= C_y'(b) \top T(\psi(b)),
\end{align*}
where
\[
C_y' = C_y \odot^{\top} P_{r_1}(a) \odot^{\top} P_{r_2}^{\intercal}(a).
\]

Inverse relations are naturally handled via matrix transposition.

\noindent\textbf{Case 2: $a$ connects to an existential variable $x$.}

Assuming a single edge $r(a,x)$,
\begin{align*}
    \phi(y)
    &= \exists x.\, \mu(x,C_x) \land r(a,x) \land \psi(y;x), \\
    A[\phi(y)](b)
    &= \bot^{\star}_{c \in \mathcal{E}}
    [C_x(c) \top P_r(a,c) \top T(\psi(b;c))].
\end{align*}

Let $C_x' = C_x \odot^{\top} P_r(a)$, yielding
\[
A[\phi(y)](b)
= A[\exists x.\, \mu(x,C_x') \land \psi(y;x)](b).
\]

Hence, constant nodes and their incident edges can be removed after propagating
constraints.

\subsection{Step 4: Cutting Leaf Nodes}

A node $u$ is a leaf if it is connected to exactly one other node $v$.
We show that such nodes can be eliminated efficiently.

\noindent\textbf{Case 1: $u$ and $v$ are both existential variables.}

Assume
\[
\phi(y) =
\exists x_1,x_2.\,
\mu(x_1,C_{x_1}) \land r(x_1,x_2) \land \mu(x_2,C_{x_2}) \land \psi(x_2,y).
\]

Define
\[
C^{\star} =
[(P_r \circ^{\top} C_{x_2})^{\intercal} \circ^{\top} C_{x_1}]^{\intercal}.
\]

Then
\begin{align}
    T(\phi(a))
    &= \bot^{\star}_{x_1=b,x_2=c}
    [C_{x_1}(b) \top P_r(b,c) \top C_{x_2}(c) \top T(\psi(c,a))] \\
    &= \bot^{\star}_{x_2=c}
    \{\bot^{\star}_{x_1=b}[C^{\star}(b,c) \top T(\psi(c,a))]\}.
\end{align}

If $\bot^{\star}$ is the G\"odel $t$-conorm (max), we have
\[
\max_b [C^{\star}(b,c) \top T(\psi(c,a))]
= \max_b C^{\star}(b,c) \top T(\psi(c,a)).
\]

Thus,
\[
C_{x_2}'(c) = \max_b C^{\star}(b,c),
\]
allowing removal of $x_1$.

\noindent\textbf{Case 2: $u=y$, $v=x$.}

\[
\phi(y) = \mu(y,C_y) \land \exists x.\, r(x,y) \land \psi(x).
\]

Let
\[
A^{\star} = P_r \circ^{\top} A[\psi(x)].
\]

Then
\[
A[\phi(y)](a) = C_y(a) \top \bot^{\star}_{b} A^{\star}(b,a).
\]

\noindent\textbf{Case 3: $u=x$, $v=y$.}

\[
\phi(y) = \exists x.\, \mu(x,C_x) \land r(x,y) \land \mu(y,C_y) \land \psi(y).
\]

Proceeding similarly, we obtain
\[
C_y'(a) = \max_b C_y^{\star}(b,a),
\]
where $C_y^{\star}$ is defined analogously.

After this step, all leaf nodes are removed.

\subsection{Step 5: Obtaining the Answer Vector}

After iteratively applying the above steps for acyclic queries, the query graph contains only the free
variable $y$ with formula $\mu(y,C_y)$. By definition, the answer vector is exactly $C_y$.